\documentclass[lettersize,journal]{IEEEtran}
\usepackage{amsmath,amsfonts}
\usepackage{algorithmic}
\usepackage{algorithm}
\usepackage{array}
\usepackage[caption=false,font=normalsize,labelfont=sf,textfont=sf]{subfig}
\usepackage{textcomp}
\usepackage{stfloats}
\usepackage{url}
\usepackage{verbatim}
\usepackage{graphicx}
\usepackage{cite}
\begin{document}

\title{Reactive Composition of UAV Delivery Services in Urban Environments}


\author{\IEEEauthorblockN{Woojin Lee, Babar Shahzaad, Balsam Alkouz, Athman Bouguettaya
}

\IEEEauthorblockA{School of  Computer Science,
The University of Sydney, Australia\\
wlee2926@uni.sydney.edu.au, \{babar.shahzaad, balsam.alkouz, athman.bouguettaya\}@sydney.edu.au
}
}




\maketitle

\begin{abstract}
We propose a novel failure-aware reactive UAV delivery service composition framework. A skyway network infrastructure is presented for the effective provisioning of services in urban areas. We present a formal drone delivery service model and a system architecture for reactive drone delivery services. We develop radius-based, cell density-based, and two-phased algorithms to reduce the search space and perform reactive service compositions when a service failure occurs. We conduct a set of experiments with a real drone dataset to demonstrate the effectiveness of our proposed approach. 
\end{abstract}


\section{Introduction}

\IEEEPARstart{T}{he} rapid growth of e-commerce has increased ground delivery traffic, affecting travel times \cite{doole2021constrained}. Drones (aka \textit{Unmanned Aerial Vehicles} (UAVs)) are a promising alternative for logistic operations \cite{guo2023drone}. Drones\footnote{In the rest of the paper, we will use Drone and UAV interchangeably} are an emerging form of IoT devices, flying in the sky with full network connectivity capabilities \cite{8119717}. This connectivity enables the efficient operation of autonomous drones as well as faster and safer responses to \textit{unanticipated changes} \cite{lee2022autonomous}. Drone delivery can reduce the dependency on ground-based road infrastructure by utilizing air space, lowering costs and emissions, reducing congestion, and improving road safety \cite{roca2019logistic,liu2022constraint}. Drone delivery has risen in popularity, particularly during the COVID-19 pandemic, as the demand for safe and contactless delivery has grown \cite{shahzaad2023optimizing}. Alphabet's Wing reported a 350\% month-on-month increase in the number of sign-ups for its delivery service during the COVID-19 pandemic\footnote{https://www.protocol.com/alphabet-wing-drone-delivery-coronavirus}.

Delivery drones require an organized and arranged medium to operate safely. Current drone delivery systems consider direct point-to-point delivery services from a source to a destination \cite{dorling2016vehicle}. Applications, like Alphabet's OpenSky, advise when the airspace is clear or restricted for a given flight\footnote{https://opensky.wing.com/}. We propose to generalize the point-to-point delivery using the \textit{service paradigm} to a \textit{skyway network} where drones may have stopovers before they reach their destination. In this respect, the service-oriented approaches ensure congruent and effective provisioning of drone delivery services in skyway networks \cite{alkouz2020swarm}. Building rooftops represent nodes connected to each other by flights, thus forming a skyway network. Each rooftop is considered a delivery service destination, a recharging station, or both. Drones in delivery services typically traverse the skyway network in Line of Sight (LOS) segments connecting the nodes \cite{10.1145/3460418.3479289}. We abstract a drone flying in \textit{skyway segment} as a drone delivery service. Drone delivery services are well-suited to being represented using the service paradigm since their fundamental constituents, namely functional and non-functional attributes, map to it \cite{shahzaad2019constraint}. As any other service, a drone delivery service exhibits functional and non-functional (aka \textit{Quality of Service} (QoS)) properties. A drone delivery service's \textit{functional} property is the delivery of packages from a source to a destination using drones. The \textit{non-functional} properties of a drone delivery service may include delivery cost, maximum payload, delivery time, and distance traveled.

A successful drone delivery service composition involves three primary steps. First, a drone-request allocation mechanism is required to assign the most suitable drone for a delivery request. An optimal allocation entails the maximum utilization of drones and assigning the most profitable delivery requests to a drone delivery service provider \cite{park8}. Second, an optimal path composition approach from the source to the destination nodes is needed, taking into account both intrinsic and extrinsic factors that impact drone delivery operation \cite{shahzaad2021top}. The intrinsic factors are related to the drone's intrinsic characteristics, including the payload and battery limitations. The extrinsic factors are related to the drone's operating environment and include the weather and the number of available recharging pads at any given node at a time \cite{janszen2021constraint}. Third, a robust drone delivery operation requires a failure-recovery mechanism. This mechanism would require a resilient composition approach, which would provide for \textit{reactive path rerouting}, i.e., dynamic service recomposition. The \textit{failure-recovery} aspect is the focus of this work.

A feasible delivery flight route assures the drones' safe arrival at their destination and the fulfillment of assigned delivery service requests when composing the delivery flight route. However, these planned feasible routes may not be valid anymore due to uncertainties. For example, unexpectedly adverse climatic circumstances (such as strong winds) can result in excessive battery power usage. As a result, the remaining battery runtime at a specific segment may be less than the flight mission's planned duration \cite{9380171}. A drone delivery service failure may occur when the QoS constraints of a skyway segment in a precomputed composition plan fluctuate at runtime. As a result, the drone delivery service may no longer provide the required QoS and fail. For example, service failure results in the early or late arrival of drones at a recharging station than the initial scheduled plan, thus affecting the desired QoS \cite{2021335}. Drone delivery service failures may also occur due to the abrupt declaration of temporary no-fly zones for security or regulatory reasons and obstacles that may obscure a skyway segment. As a result, the planned flight paths must be dynamically rerouted to ensure that packages are delivered safely and successfully. 

Most of the research done on drone delivery services focuses on proactive planning and optimizing flight routes. These approaches primarily function in offline modes prior to the drones' flight \cite{alkouz2020swarm}. Routes chosen by proactive planning ensure that the planned mission's purpose is attained while considering any foreseeable variables. The drawback of these approaches is their inability to reroute drones when an unforeseeable circumstance occurs during the flight. Few works on the reactive rerouting of a drone's path focus solely on the drones' safe return to their sources \cite{grzegorz2021reactive,kim2020real}. However, these solutions are not feasible for the time-constrained drone delivery service requests. Therefore, a reactive UAV delivery service composition approach is required to provide a quick response to drone delivery service failures caused by unforeseen dynamic circumstances.

We propose a \textit{failure-aware reactive UAV delivery service composition framework} that reduces the search space and finds a path to connect the disconnected segments in the precomposed composition plan. In this context, the \textit{global} and \textit{local} compositions are the two principal approaches for drone rerouting. In a global composition, when a failure happens at a segment, the new path to the destination is computed considering all the nodes in the skyway network \cite{10.1007/978-3-662-45391-9_26}. In a local composition, a new path that connects the two disconnected nodes is computed. This approach usually uses a subarea of the entire skyway network. The benefit of this approach is reducing the search space, which is reflected in the reduced computational times. Our proposed framework includes three reactive service composition algorithms to address the failures for effectively provisioning drone delivery services. Each algorithm constrains the search space using geometric methods and finds an optimal composition plan for the failed delivery services. We summarize our main contributions as follows:
\begin{itemize}
    \item  A novel failure-aware reactive UAV delivery service composition framework in the event of failures
    \item Three reactive service composition algorithms for local optimal reactive service compositions of UAV delivery services
    \item A complete assessment of the proposed algorithms in comparison to the global composition approach using a real drone dataset to demonstrate the performance of the proposed composition approach
\end{itemize}

\subsection{Motivating Scenario}
We use a UAV-based package delivery scenario as our motivating scenario. Suppose a company offers drone-based solutions for package delivery in Texas, USA. The company plans to deliver a package from \textit{San Marcos} to \textit{San Antonio} (94 km). The range of a typical delivery drone varies from 3 to 33 km \cite{alkouz2021service}. The drone flight range is further affected by its speed and payload weight \cite{shahzaad2021robust}. Drone service failures may occur during the delivery operation due to extrinsic factors (e.g., weather conditions) and fluctuating QoS parameters. In such a scenario, it is crucial to address the factors mentioned above, especially the drone service failures for time-optimal and successful delivery of packages.

We construct a skyway network to provide conflict-free routes to UAVs. In this respect, a conflict-free route follows the Federal Aviation Administration's drone flying regulations, such as flying within the LOS and avoiding flying over no-fly zones. A skyway network enables the safe and scalable deployment of drone-based delivery solutions in shared airspace \cite{shahzaad2022drone}. We abstract each line segment in the skyway network as a service that is served by a drone. Each skyway network node is a fixed landing pad on the rooftop of a high-rise building within the Texas area. Each node may simultaneously serve as a \textit{delivery target} or a \textit{recharging station}. UAV delivery services may fail due to uncertain conditions. For example, a drone service may arrive late (or early) due to a strong headwind (or tailwind) or may not find a recharging pad available at a particular recharging station due to recharging constraints and stochastic arrival of other drone services. As a result, the drone service may no longer provide the required QoS and fail. We reformulate the drone delivery problem as a reactive service composition problem to address the failures in drone delivery services and ensure the successful delivery of packages to the desired destinations.

\section{Related Work}
This paper combines concepts from two separate areas: (1) drone delivery services and (2) failures and recovery in composite services. In this section, we overview related work in these two areas.

\subsection{Drone Delivery Services}
In recent years, several approaches have been proposed for the effective provisioning of drone delivery services \cite{shahzaad2019composing,shahzaad2020game}. An energy model for drone delivery services is presented considering the nonlinear behavior of drone battery consumption \cite{10.1145/3218603.3218614}. The proposed model incorporates three main factors that influence the battery consumption of a drone: travel distance, speed, and payload weight. It is indicated that the battery consumption of a drone grows linearly as the distance increases. However, the battery consumption grows exponentially with the increasing payload weight and drone speed. A battery-aware drone delivery scheduling algorithm is proposed to serve more delivery requests with the same battery capacity. The drone performs the single package delivery in the proposed algorithm and returns to the depot. The proposed model does not consider the drone flight regulations, such as flying within the LOS and avoiding no-fly zones to provide delivery services. In addition, the proposed energy model does not incorporate the effects of environmental uncertainties on drone deliveries that may cause failures.

Energy minimizing and range-constrained drone delivery problem is studied where deliveries are made solely using drones \cite{DUKKANCI2021102985}. In this study, a ground vehicle serves as a mobile station for the take-off and landing of the delivery drones. The objective of this study is to minimize the operational cost and energy consumption of drones. A nonlinear model of the drone delivery problem is presented and reformulated as a second-order cone program to detect the energy consumption patterns of drones. The perspective cuts are used to strengthen the cone program, which is then solved using off-the-shelf commercial solvers. It is concluded that a drone's flight range and energy consumption depend on its speed. The proposed drone delivery problem does not capture the real-world scenarios where drone delivery services are affected by the inherent limitations of drones and environmental uncertainties. These inherent limitations and environmental uncertainties may trigger failures in the delivery of services.

A collision-free trajectory scheduling system is proposed considering the energy constraints of a drone \cite{10.1007/978-3-030-64843-5_9}. This study aims to maximize the drone flight time by landing and swapping batteries along the delivery path at charging stations. Each drone visits a set of predefined warehouses to accomplish the parcel delivery task. A multi-source A* algorithm is proposed to compute the scheduling of each drone. The drones fly at a distance from each other to avoid collisions. The proposed scheduling system is scalable to large metropolitan networks because the proposed algorithm's time complexity is polynomial. The simulation experiments demonstrate the proposed approach's efficiency and scalability with the network size and number of delivery requests. It is observed that the computation time is an increasing linear function of the number of charging stations, i.e., the computation time increases as the number of charging stations increases. \textit{The proposed scheduling system does not consider the congestion conditions at recharging stations, drone flying regulations, and failures caused by environmental uncertainties such as wind.}

A modular optimization method is proposed to increase the drone fleet readiness and decrease the fleet size for deliveries \cite{7934790}. A module in the optimization method lends more flexibility to drone operations with its interchangeable components. The components include replaceable batteries, motors, carriers, and propellers. A forward-looking strategy is used to increase the performance of drone deliveries. The proposed approach compares modular delivery drones with non-modular delivery drones. The simulation experiments illustrate the modular optimization method's effectiveness in minimizing power consumption and drone delivery time. \textit{The proposed optimization method does not consider the impact of environmental uncertainties such as wind that may cause failures and congestion conditions caused by other drones in the same delivery system}.

A drone service system is presented to provide long-distance delivery services considering refueling and maintenance of drones \cite{9099809}. This system aims to minimize the travel distance of a drone and the number of landing depots during the delivery operation. An ant colony algorithm with the A* algorithm is proposed to solve the problem of long-distance delivery services. \textit{The proposed system does not consider the factors affecting a drone's flight range, such as payload and wind conditions}. The proposed drone delivery system also does not consider drone service failure.

A single on-drone-decision system is modeled as a multi-agent system to enhance drone service delivery \cite{alwateer2021drones}. The agents in this system are clients (service receivers) and drones (owned by service providers). The factors affecting the provisioning of drone services are also identified. This study aims to investigate the trade-offs between maximizing profit and client satisfaction. In addition, drone behavior is examined for various factors, including drone speed, service duration, and distribution of clients. The model's generality is assessed using stereotypical distributions to cover a wide range of scenarios. It is indicated that the affordability of a service is the most critical factor that influences the utilization of drones-as-a-service. The simulation experiments are performed to evaluate the impact of drone decision-making strategies on clients and service providers. It is concluded that battery constraints influence the performance of strategies. \textit{The proposed approach does not capture a range of factors that affect drone delivery services. These factors include drone flying regulations and failures caused by environmental uncertainties such as wind speed and wind direction.}

A drone service framework is proposed to provide delivery services \cite{9380171}. The proposed framework includes a service model for drone services based on the spatio-temporal attributes of drones. Scheduling, route planning, and composition are the fundamental components of the proposed approach. The scheduling generates itineraries for drones in a skyway network. A route-planning algorithm is proposed, focusing on the selection of the optimal route. The drone services are composed using a drone service composition algorithm at each station. \textit{This framework does not consider the recharging requirements of drones and congestion conditions at stations.} In addition, the failures in drone delivery services are not modeled in the proposed drone service framework that are required to ensure persistent service delivery.

\subsection{Failures and Recovery in Composite Services}

A recovery approach is presented to recover failures in composite services \cite{Saboohi:2012:FRW:2428736.2428787}. The proposed approach is divided into two phases: the offline phase and the online phase. The subdigraphs of services are generated and added to the composite services in the offline phase. The subdigraphs are pre-calculated to speed up the replacement process in this phase. The composite services are executed in the online phase. The semantic description of composite services is used to rank the subdigraphs. A failure recovery algorithm is designed for the online phase that includes forward and backward approaches. The forward approach tries to re-execute the failed services to achieve the desired objective of composite service. The backward approach is used when the forward approach is unsuccessful. The proposed approach does not consider the QoS features of services and the dynamic environment in which the services operate.

A constraint-aware failure recovery approach is proposed to predict failures in the composite services \cite{LALEH2018387}. The proposed approach initially predicts and verifies failures in a composite service for the correct execution of services. The objective of this study is to minimize the number of service rollbacks upon service failures. The proposed failure recovery approach comprises a planning-based algorithm for initial service composition, and a constraint-processing method proceeds with the service composition for the successful delivery of services. The planning-based algorithm composes the services based on the service composition request. The constraint-processing method first verifies the constraints in the composite service. This method is then used to predict and recover failures in the composite service. The proposed failure recovery approach is limited to generating solutions for a small number of services. This approach does not incorporate the dynamism of the service environment for the service composition and recovery of failures.

A service reconfiguration approach is presented to repair the failed services and meet the end-to-end requirements of users \cite{5175901}. The proposed approach is based on reconfiguration regions that consist of failed services. The reconfiguration regions of failed services are merged if they overlap. The new services replace the failed services at runtime. A reconfiguration region identification algorithm is proposed to identify the faulty regions. The identified faulty regions are recomposed using the specified QoS constraints to obtain the desired composition. The extended constraints are used to relax the original constraints if new sub-processes for the reconfiguration regions do not exist.

A drone delivery scheduling problem is proposed, taking into account the reliability of drones \cite{8453380}. This study aims to minimize the expected loss of demand that unsatisfied customers represent. Customer satisfaction is mainly affected by drone failures during the delivery operation. It is assumed that drone failures occur in an exponential distribution manner. A two-stage stochastic programming approach is proposed to solve the drone scheduling problem. The first stage of the proposed approach provides a solution pool of feasible paths for the scheduling problem. An algorithm is presented to calculate the expected loss of demand for each path in the solution pool. The second stage uses the results of the first stage to provide the most reliable delivery schedule covering all customers. The proposed drone delivery scheduling problem considers the failures caused by environmental uncertainties. However, the drone flying regulations, recharging requirements of drones, and congestion conditions at recharging stations are not considered.

A resilient composition framework for drone delivery services is proposed considering congestion conditions at recharging stations \cite{2021335}. The framework includes a formal service model for representing constraint-aware drone services. A deterministic lookahead algorithm generates an initial offline drone service composition plan. A heuristic-based resilient composition approach is proposed to adapt to the runtime changes in the initial composition plan and update it to meet the delivery requirements of the user. This paper defines a failure as the drones' early or late arrival at certain recharging stations. \textit{The proposed resilient composition framework considers only the deterministic failures in drone services}.

The path planning problem for scheduling drone delivery services is formulated as a constraint satisfaction problem \cite{8453346}. This study aims to find the shortest, flyable, and collision-free paths for service delivery in dynamic graphs. A collision-free path planning algorithm is designed to solve the drone delivery problem. The proposed algorithm initially finds the shortest path to the destination using a well-known A* algorithm. Then, it checks if the drone can fly to the respective destination, considering the drone's flight range. Finally, the algorithm checks if the selected path is flyable for the drone at the scheduled time. In this context, a flyable path represents a path where the drone does not collide with other drones or stationary obstacles. The drone is then scheduled to traverse a path if the path is flyable. It is concluded that the increasing number of requests decreases the number of successfully scheduled paths. The proposed path planning problem focuses on avoiding drone failures by selecting collision-free paths. However, the proposed approach does not consider the drone flying regulations, recharging requirements of drones, and drone failures that occur due to environmental uncertainties. To the best of our knowledge, this paper is the first attempt to present a failure-aware reactive UAV delivery service composition framework that addresses the failures in drone delivery services.

\section{Failure-Aware Reactive UAV Delivery Service Composition Framework}

We present a failure-aware reactive UAV delivery service composition framework to ensure the successful delivery of packages using drones. The proposed framework includes (1) Skyway Network Infrastructure, (2) Drone Delivery Service Model, (3) System Architecture for Reactive Drone Delivery Services, and (4) Reactive Service Composition Algorithms for Drone Delivery Services.

\subsection{Skyway Network Infrastructure}

A skyway network infrastructure enables the realization of multi-point UAV-based delivery solutions \cite{bradley2023service}. We construct a skyway network by joining a set of predefined line segments of which endpoints are network nodes. Each line segment in the skyway network satisfies the LOS drone flying requirements. Furthermore, the line segments avoid the no-fly zones and restricted areas. Each skyway network node represents a rooftop of a high-rise building. Each node may act both as a delivery target and/or a recharging station to support the persistent delivery of drone services. Our proposed skyway network uses a city's existing building rooftop infrastructure to minimize the deployment cost of recharging stations. We formally model the skyway network infrastructure as an undirected graph $G = (N, E)$, where $N$ is a set of nodes and $E$ is a set of edges. Each node (i.e., recharging station) has a \textit{finite number of pads}. Each edge represents a line segment service that is served by a drone. Suppose $D = \{d_1, d_2,\ldots, d_n\}$ is a set of drones that operate in the same skyway network. In this regard, $B$ is a set of battery capacities for all drones in $D$. The delivery cost and amount of battery required to transport a package from a node $i$ to node $j$ are represented by $c_{ij}$ and $b_{ij}$, respectively. The payload weight, traveling distance, and travel speed are directly proportional to a drone's battery consumption.

\subsection{Drone Delivery Service Model}

We formally model the drone delivery service, customer delivery request, drone delivery service failure, and reactive service composition problem as follows.

\textbf{Definition 1: Drone Delivery Service}. A \textit{drone delivery service} is a tuple $<DDS_{id},$ $DDS_{f}, DDS_q>$, where
\begin{itemize}
    \item[$\bullet$] $DDS_{id}$ is a unique drone delivery service ID,
    \item[$\bullet$] $DDS_{f}$ is the delivery function of a drone $d_i \in D$ to transport a package over a line segment. The location and time of a drone delivery service are tuples $<loc_s,loc_e>$ and $<t_s, t_e>$, where
    \begin{itemize}
        \item $loc_s$ and $loc_e$ are the start and the end locations of a drone delivery service,
        \item $t_s$ and $t_e$ are the start and the end times of a drone delivery service,
    \end{itemize}
    \item[$\bullet$] $DDS_q$ is a set of QoS attributes of a drone delivery service represented by a tuple of $<q_1, q_2,\ldots,q_n>$. Examples of quality attributes include payload capacity, battery capacity, flight time, etc.
\end{itemize}

\textbf{Definition 2: Customer Delivery Request}. A \textit{customer delivery request} is a tuple $<\zeta, \xi, rt_{s}, w>$, where
\begin{itemize}
    \item[$\bullet$] $\zeta$ represents the source (i.e., warehouse),
    \item[$\bullet$] $\xi$ represents the desired destination (i.e., customer location),
    \item[$\bullet$] $rt_s$ represents the start time of the request,
    \item[$\bullet$] $w$ represents the weight of the package to be delivered.
\end{itemize}

\textbf{Definition 3: Drone Delivery Service Failure}. A \textit{drone delivery service failure} is defined as the deviation of a drone delivery service from its expected behavior. A drone delivery service failure is represented as a tuple $<FT, DDS_i, LOC, TS>$, where 
\begin{itemize}
    \item[$\bullet$] $FT$ represents the type of failure. The types of failures include environmental (adverse weather impacts), operational (battery, payload limitations), navigational (unexpected obstacles), regulatory (imposition of no-fly zones or airspace restrictions that affect the initially composed skyway path), infrastructure (unavailability of recharging pads at particular recharging stations), and service-level (precomputed QoS parameters are no longer met due to uncertain conditions) failures, leading to early or late arrivals of drones at intermediate stations or deviations from the scheduled plan,
    \item[$\bullet$] $DDS_i$ represents the failed drone delivery service,
    \item[$\bullet$] $LOC$ represents the location of failed service,
    \item[$\bullet$] $TS$ represents the timestamp of failed service.
\end{itemize}

\subsection{Problem Formulation}
Given a set of drone delivery services \( S_{DDS} = \{DDS_1, DDS_2, ..., DDS_n\} \) and a customer delivery request \( <\zeta, \xi, rt_s, w> \), we formulate the drone delivery problem as a reactive service composition problem of drone delivery services. This problem focuses on selecting and composing optimal skyway segments (i.e., drone delivery services) in the event of service failures. The term \textit{reactive} means responding to the variations in the precomputed service composition plan. The types of failures considered primarily include the unavailability of skyway segments. These disruptions necessitate the recomposition of the drone's skyway path to ensure successful package delivery from the source \( \zeta \) to the destination \( \xi \). The problem is constrained by intrinsic factors such as the drone's payload \( w \) and battery life \( B \), as well as extrinsic factors such as changing weather conditions and airspace regulations such as the emergence of temporary no-fly zones. These constraints are critical as they affect the availability and selection of skyway segments, posing a challenge in meeting a customer's delivery request's QoS requirements. The key challenges include the ability to dynamically adjust to sudden changes in the operational environment, such as disruptions in precomputed paths due to uncertain service failures and constraining the search space to expedite the drone service recomposition process. The objective of this formulation is to optimize the drone service composition process under these constraints and challenges. This involves minimizing the travel distance while ensuring a rapid recomposition of the skyway path in case of failure.

\subsection{System Architecture for Reactive Drone Delivery Services}

\begin{figure} [t]
    \centering
    \includegraphics[width=0.9\linewidth]{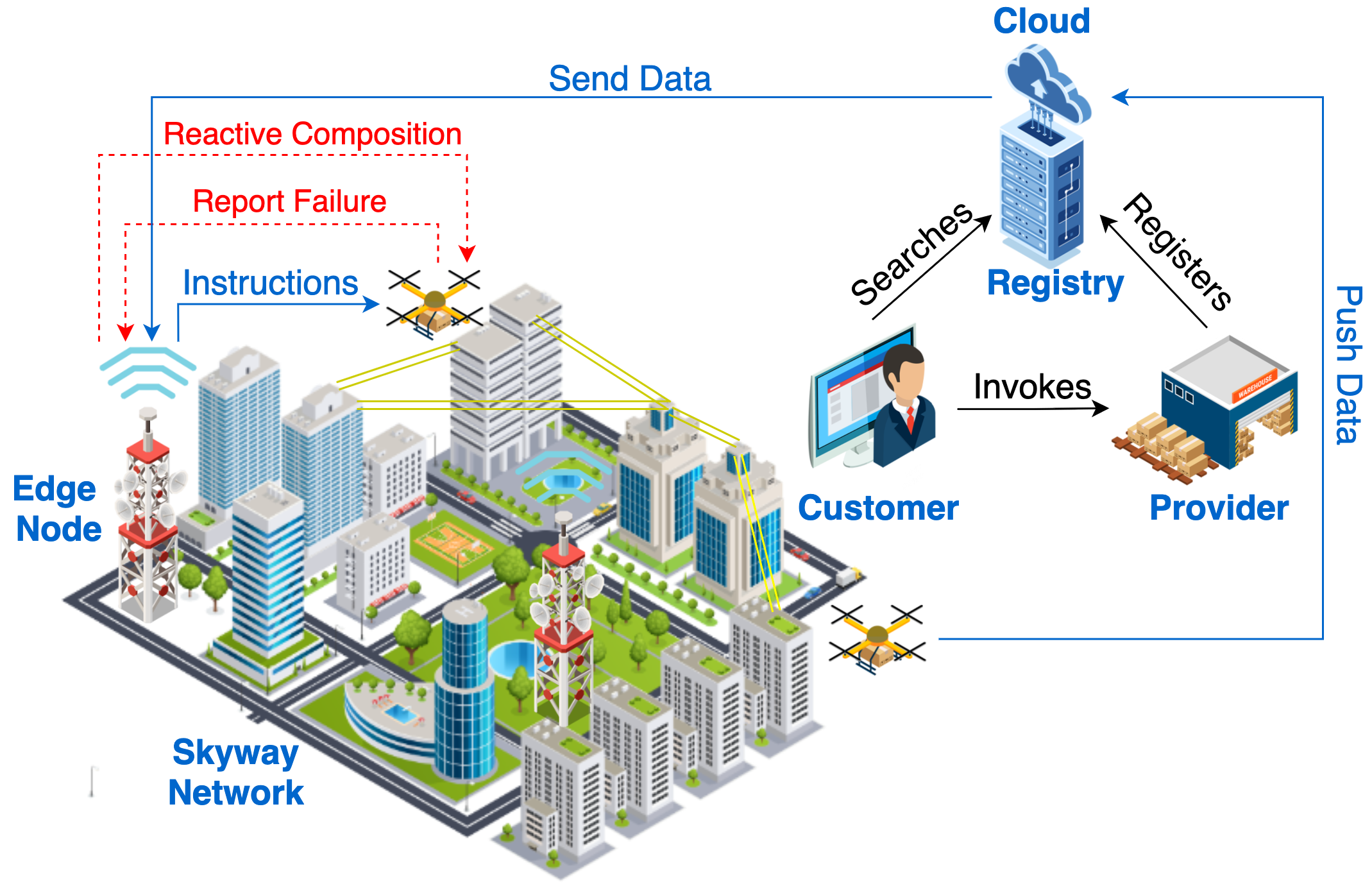}
    \caption{System Architecture for Reactive Drone Delivery Services}
    \label{Architecture}
\end{figure}

We present a high-level system architecture design for reactive drone delivery services. The design is based on drones delivering packages from a single source to a single destination with being prone to potential failures. The architecture depicted in Fig. \ref{Architecture} is an adapted service-oriented architecture with a sensor-cloud infrastructure. To publicize their services, the \textit{providers} list them in the registry. The customer searches the registry for services and invokes preferred services. When a customer invokes a service, the drone delivery management system initially computes the most efficient path composition. Drones, edge nodes, and the cloud are expected to share computational loads. At the \textit{drone} level, simple computations are performed. At the \textit{edge}, more complex computations are executed. Edge nodes are often distributed in strategic locations to assist delivery \cite{shahzaad2022service}. The \textit{cloud} is used for computations that require a lot of data. As the drones move, they send their locations and battery levels to the cloud, which houses all the drones' and requests' data. When needed, the system at the edge nodes performs rapid computations, especially in the event of a failure. When a drone detects a failure at a skyway segment, it sends a direct message to the edge, as shown in Fig. \ref{Architecture}. In this situation, the edge reactively computes a new path composition to bypass the failed skyway segment and securely direct the drone to its destination. The drone function is a dichotomy between informing the cloud of its location and receiving instructions from the edge on where to go. When the drone arrives at its destination, the customer is notified of the successful package delivery.

\subsection{Reactive Service Composition Algorithms for Drone Delivery Services}

We propose three reactive service composition algorithms for the effective provisioning of drone delivery services. Each algorithm focuses on reducing the search space to enhance its computational complexity. The proposed reactive service composition algorithms are (1) Radius-Based Reactive Service Composition, (2) Cell Density-Based Reactive Service Composition, and (3) Two-Phased Reactive Service Composition.

\subsubsection{Radius-Based Reactive Service Composition}

When a failure occurs at a skyway segment connecting two nodes, e.g., A and B, a local reactive service composition tries to find an alternative path to B from A. To \textit{efficiently} compose a path to B, the search space should be narrowed to include nodes that will potentially be used to form the new path. In this approach, a fixed-shaped bounding area, i.e., a circle, is computed to encompass the two disconnected nodes. Let $N$ be the set of skyway network nodes that lie within the bounding circle area, $N$ = \{{$n_1$, $n_2$,..., $n_m$}\}, and $m$ is the number of nodes encompassed within the circle area. In this case, the local optimal search algorithm would only consider the nodes in N when composing a new path. The size of the bounding area in this approach is proportional to the length of the broken segment, creating a larger bounding area for longer segments. The most significant advantage of this approach is that it only requires the position of two disconnected nodes to construct the bounding area. As a result, the computing time for the preprocessing phase, prior to the local optimal search within the bounding area, is reduced.

\begin{algorithm}[t]
\caption{Radius-Based Reactive Service Composition}
\begin{flushleft}
\hspace*{\algorithmicindent} \textbf{Input: Network, Node A, Node B} \\
\hspace*{\algorithmicindent} \textbf{Output: Path} 
\end{flushleft}
\begin{algorithmic}[1]
\STATE $dist = \sqrt{(A.x - B.x)^2 + (A.y - B.y)^2}$
\STATE path = None
\WHILE{path == None} 
    \STATE {$B_a$ = Bounding area predefined as a shape with the radius \textit{dist}}
    \STATE {path = path from Node A to B within \textit{$B_a$}}
    \IF{path != None}
        \STATE break
    \ENDIF
    \STATE $dist$ += Network size $\times$ 0.2
    \IF{$dist$ $>$ Network size $\times$ 0.5}
        \STATE Use entire network to perform global optimal search
        \STATE break
    \ENDIF
\ENDWHILE
\RETURN path

\end{algorithmic}
\label{radius-based}
\end{algorithm}

Algorithm \ref{radius-based} describes the main steps involved in the radius-based reactive service composition approach. The algorithm takes the entire network as an input along with the two disconnected nodes, A and B. The algorithm returns the recomposed path between nodes A and B. The new path is exclusively composed of nodes that are located within the computed bounding area. In the beginning, the Euclidean distance ($dist$) between points (i.e., nodes) A and B is calculated (Line 1). This distance is used as the radius to draw a circle at the center point of the broken segment (Line 4). This circle encompasses nodes A and B and other nodes within the bounded region. The local optimal path search between A and B is performed only in the bounding area. The path is returned if it exists within the bounding area. Otherwise, the circle's radius is incremented by a fixed value, and the local search is performed again. If the radius becomes greater than half of the network size (Line 10), the bounding area becomes as large as the entire network. Therefore, the global optimal search is used instead, where all the nodes within the network are considered when searching for a new path between nodes A and B. Although the preprocessing steps of this approach are time-efficient, it has considerable drawbacks. First, the performance is inconsistent due to skyway networks' varied layouts and node distributions. Second, the size of the bounding area is not intuitive. It is based on a fixed value and does not cater to the network's shape.

\subsubsection{Cell Density-Based Reactive Service Composition}

\begin{figure} [t]
    \centering
    \includegraphics[width=0.6\linewidth]{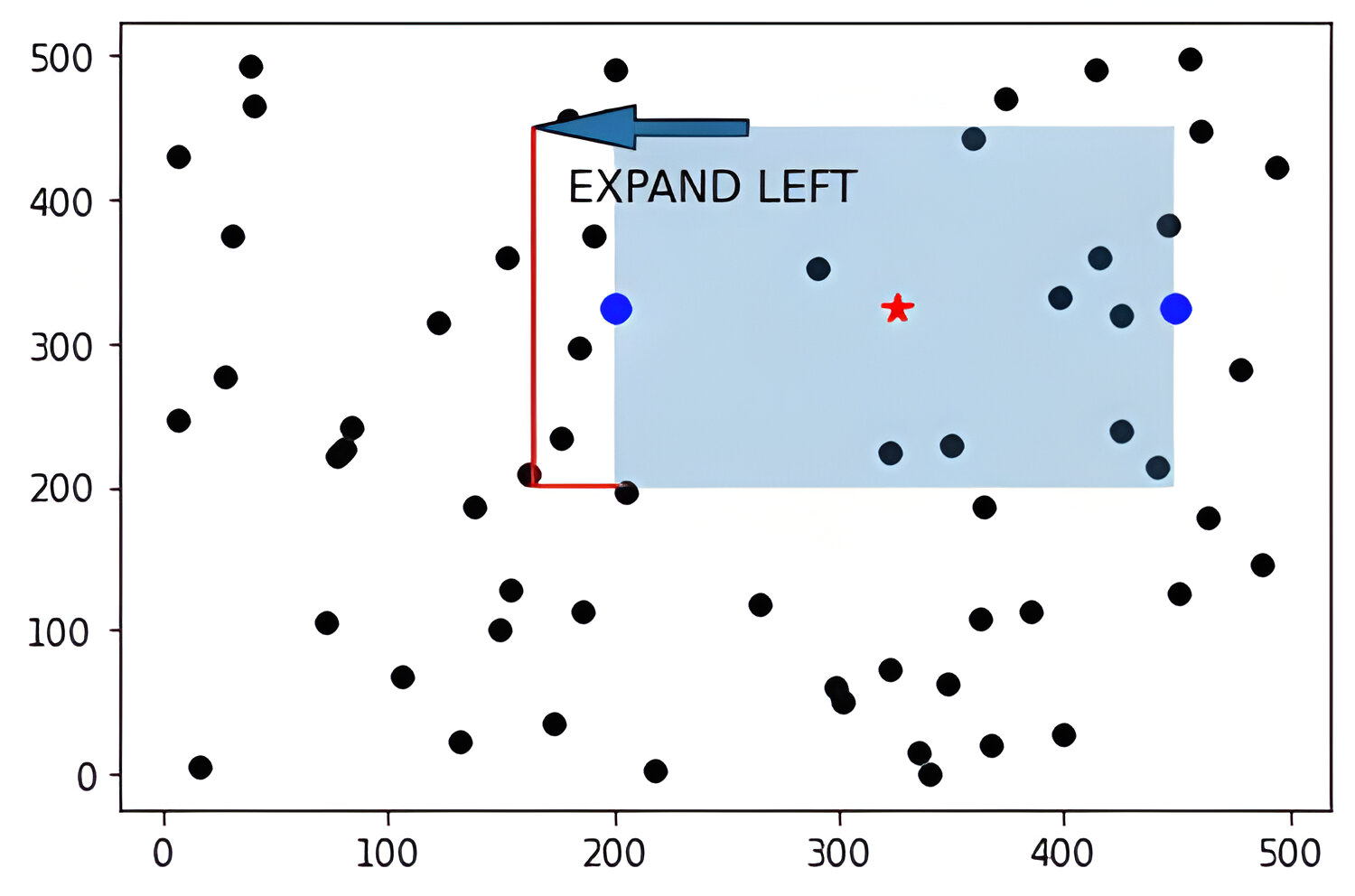}
    \caption{Skyway Network Node Distribution}
    \label{node_dist}
\end{figure}

We propose a cell density-based reactive service composition approach to overcome the identified limitations of the radius-based approach. In this approach, we consider two types of attributes. The first is the \textit{heuristic attribute}, which represents the density of the nodes. The second is the \textit{deterministic attribute}, which represents the neighbors of the disconnected nodes. Typically, the nodes in a skyway network are placed disproportionately over the map. Therefore, the number of nodes at different parts of the bounding area varies when a bounding area is defined at equal distances to all directions surrounding the disconnected nodes. Hence, the density of nodes at different parts of the network is considered to maintain a similar number of nodes in all directions from the disconnected nodes. This consideration allows us to determine whether the bounding area should be expanded more or less in a particular direction. Fig. \ref{node_dist} represents a possible network distribution of nodes in a skyway network. As shown in Fig. \ref{node_dist}, when expanding a search space bounding area, the density of the neighboring area is considered. The search space area is extended when no path connecting the nodes is found within the current bounding area. Therefore, in this case, the bounding search space is expanded to the left depending on the density of the nodes in the neighboring area. 

In this approach, when constructing a path between two different nodes, A and B, the path must contain at least one neighboring node of A and one neighboring node of B. This approach creates a bounding area that contains all the neighboring nodes of A and B, which becomes the minimum size of the bounding area. Then, concerning the heuristic attribute, i.e., the density of nodes, the different parts of the bounding area are expanded in different sizes, which get merged together.

\begin{algorithm}[t]
\caption{Cell Density-Based Reactive Service Composition}
\begin{flushleft}
\hspace*{\algorithmicindent} \textbf{Input: Network, Node A, Node B, Cell Size C} \\
\hspace*{\algorithmicindent} \textbf{Output: Path} 
\end{flushleft}
\begin{algorithmic}[1]

\STATE $Cells = [\frac{Network Height}{C}][\frac{Network Width}{C}]$
\STATE $C_{ij}$ = Nodes in Cells[i][j]
\STATE $CO = Max(num(C_{ij})) - Min(num(C_{ij}))$
\STATE path = None
\STATE NN = Set of neighboring nodes of A and B
\STATE DO = Size(Cells[i][j])

\item[]
\textbf{Phase 1:}

\FOR{Each Cell $\in$ Cells}
    \IF{0 $\leq$ num(Cell) $\leq$ $\frac{1}{3}$$\times{CO}$}
        \STATE Cell = Sparse of Nodes
    \ELSIF{$\frac{1}{3}$$\times{CO}$ $\leq$ num(Cell) $\leq$ $\frac{2}{3}$$\times{CO}$}
        \STATE Cell = Average of Nodes
        
    \ELSE
        \STATE Cell = Dense of Nodes
    \ENDIF
\ENDFOR

\item[]
\textbf{Phase 2:}
\STATE \textit{$density$}: \{DENSE = 1, AVERAGE = 2, SPARSE = 3\}
\WHILE{path == None}
    \FOR{Cell which node n $\in$ NN belongs to}
        \STATE Partial bounding area = [n.y - \textit{$density$} $\times$ DO: n.y + \textit{$density$} $\times$ DO]
        [n.x - \textit{$density$} $\times$ DO: n.x + \textit{$density$} $\times$ DO]
    \ENDFOR
    
    \STATE $B_a$ = Sum(Partial bounding areas)
    \STATE {path = path from Node A to B within \textit{$B_a$}}
    \IF{path != None or $B_a$ == Network}
        \STATE break
    \ENDIF
    \STATE Increment DO
\ENDWHILE
\RETURN path
\end{algorithmic}
\label{density-based}
\end{algorithm}

\begin{figure} [t]
    \centering
    \includegraphics[width=0.6\linewidth]{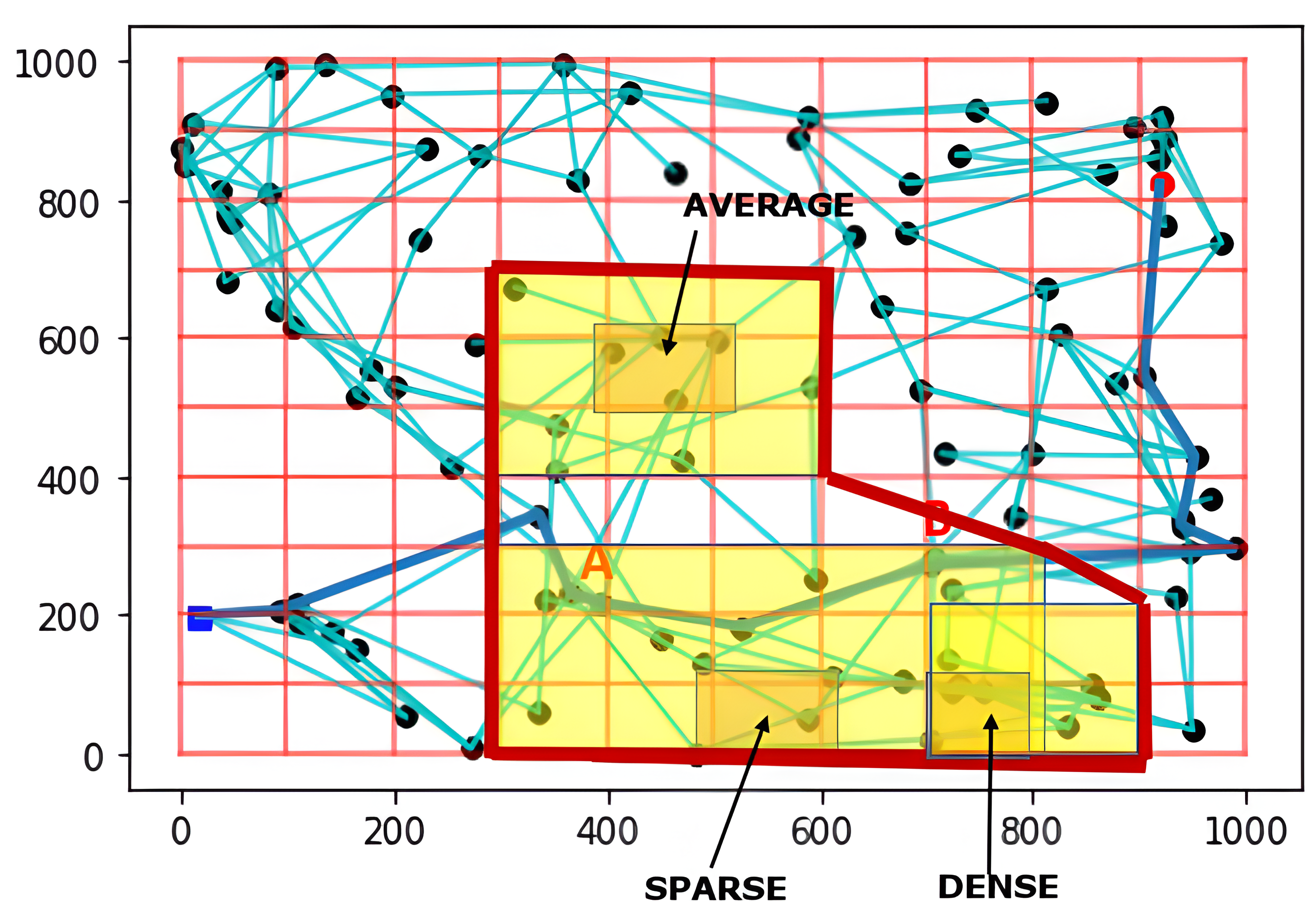}
    \caption{Cell Density-Based Reactive Service Composition}
    \label{density_based_sample}
\end{figure}

Algorithm \ref{density-based} depicts the cell density-based approach. The algorithm starts by dividing the network into smaller cells where the number of cells in the network = $\frac{Height_{size}}{Cell_{size}}$ $\times$ $\frac{Width_{size}}{Cell_{size}}$. The number of nodes within each cell determines whether the cell is \textit{sparse, average, or dense} (Lines 8-14). The algorithm uses the neighboring nodes of A and B and examines which cells they belong to. The cells are checked for the denseness of the nodes and are allocated different sizes of \textit{partial bounding areas}. A sparse cell would have a larger partial bounding area than a dense cell to cover more nodes (Lines 18-23). Then, the partial bounding areas are merged to create the final bounding area $B_a$ and perform a local optimal path search within the nodes in $B_a$. Fig. \ref{density_based_sample} illustrates how the bounding area is created. The orange cells represent the location of neighboring nodes. The yellow cells represent the partial bounding areas. The cells bounded by the red lines represent the final bounding area $B_a$. If the path is not found in $B_a$, the algorithm reallocates larger partial bounding areas for each cell and repeats the local optimal search. The iteration terminates when a path is found or $B_a$ equals the full network size.

\subsubsection{Two-Phased Reactive Service Composition}
The radius-based and cell density-based approaches consider bounding areas, which include nodes that are not particularly close. These nodes are less likely to be part of the new composed path. We propose a two-phased reactive service composition approach to overcome this issue and further restrict the bounding area. We assume that the new local optimal path most likely lies between the disconnected nodes in this approach. Farther nodes are less likely to be used as they yield longer paths in the path composition.

\begin{algorithm}[t]
\caption{Two-Phased Reactive Service Composition}
\begin{flushleft}
\hspace*{\algorithmicindent} \textbf{Input: Network, Node A, Node B} \\
\hspace*{\algorithmicindent} \textbf{Output: Path} 
\end{flushleft}
\begin{algorithmic}[1]
\STATE $L_1$, $L_2$, $L_3$, $L_4$, $L_5$ = Eq. (\ref{eq:2}), Eq. (\ref{eq:3}), Eq. (\ref{eq:4})

\STATE $R_{ba}$ = Area surrounded by ($L_2$, $L_3$, $L_4$, $L_5$)
\STATE $B_{a1}$, $B_{a2}$ = $R_{ba}$ divided by $L_1$
\STATE $B_{a}$ = max($B_{a1}$, $B_{a2}$)
\STATE $B_{a3}$ = Rhombus created by connecting the mid-points of ($L_2$, $L_3$, $L_4$, $L_5$)
\STATE $B_{a}$ = Overlapping area of ($B_{a}$ and $B_{a3}$)

\item[]
\textbf{Phase 1:}
\STATE $path$ = None
\STATE $Area$ = [$(B_{a}, 0.25)$, $(B_{a3}, 0.5)$, $(R_{ba}, 1)$]
\FOR{$a \in Area$}
    \IF {$num(nodes \in a[0]) \leq a[1]$}
        \STATE continue
    \ENDIF
    \IF{$path$ found in \textit{a}}
        \STATE break
    \ENDIF
\ENDFOR

\item[]
\textbf{Phase 2:}

\WHILE{path == None}
    \FOR{$n \in R_{ba}$}
        \STATE $R_{ba}$.add(n.closest neighbor)
    \ENDFOR
    
    \IF{number of nodes($R_{ba}) \geq$ 0.5$\times${number of nodes(network)}}
        \STATE path = global optimal path search on network
        \STATE break
    \ELSE
        \STATE path = local optimal path search on $R_{ba}$
    \ENDIF
\ENDWHILE

\RETURN path
\end{algorithmic}
\label{two-phased}
\end{algorithm}

\begin{figure} [t]
    \centering
    \includegraphics[width=0.6\linewidth]{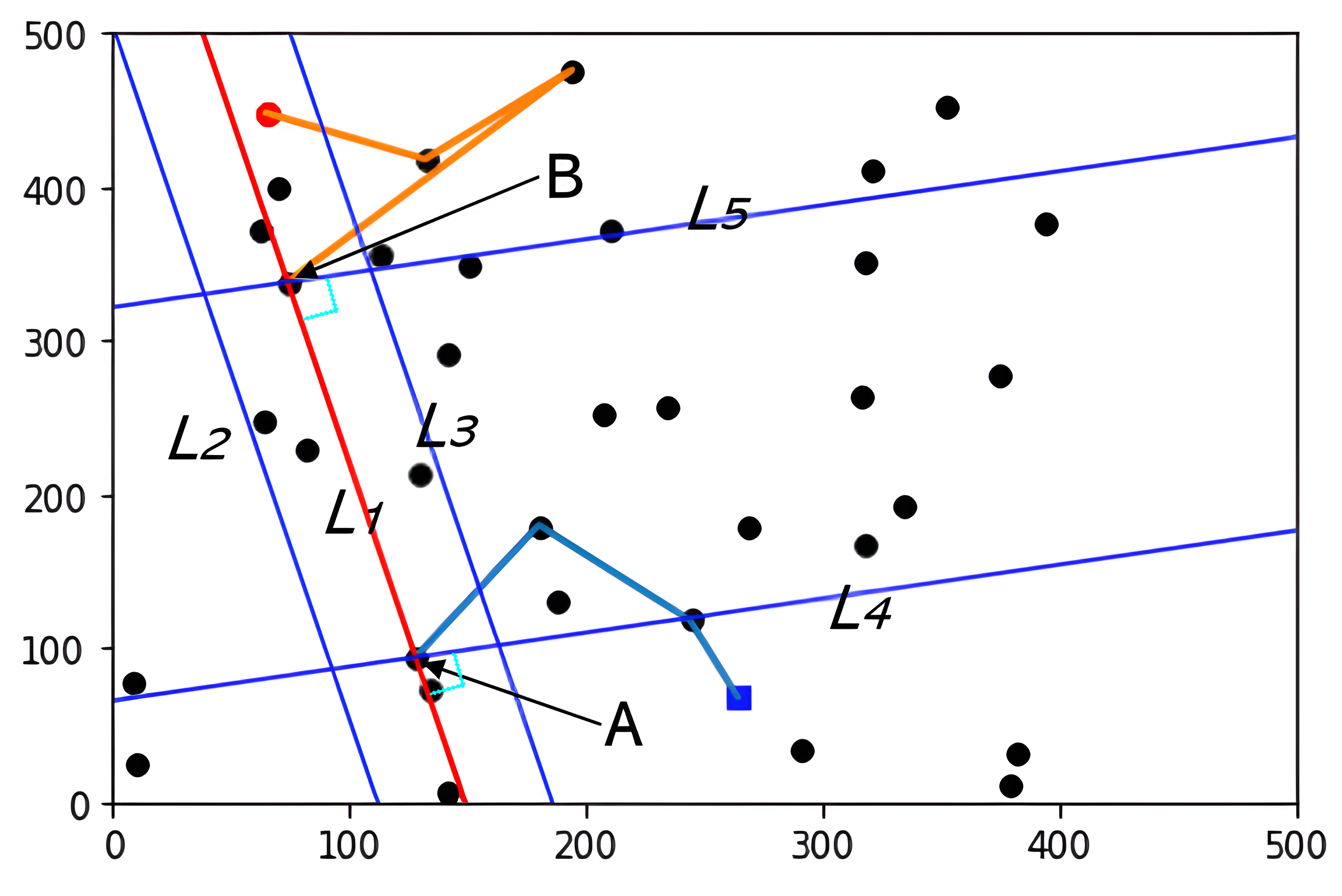}
    \caption{Rhombus Bounding Area Illustration}
    \label{Rhombus}
\end{figure}

Algorithm \ref{two-phased} describes the two-phased reactive service composition approach. Before executing any phase, a rhombus-shaped bounding area between the disconnected nodes, e.g., A and B, is created. Let a linear line, $L_1$, that passes through the disconnected nodes A and B be created using Eq. (\ref{eq:2}). Two lines, $L_2$ and $L_3$, are created parallel to $L_1$ by adding and subtracting specified value to c using Eq. (\ref{eq:3}). Then, two lines, $L_4$ and $L_5$, are created perpendicular to $L_1$ using Eq. (\ref{eq:4}). The area surrounded by $L_2$, $L_3$, $L_4$, and $L_5$ creates a rhombus-shaped bounding area, $R_{ba}$, between A and B. The blue lines in Fig. \ref{Rhombus} show the rhombus bounding area created.

\begin{equation}\label{eq:2}
    L_1: y = mx + c
\end{equation}
\begin{equation}\label{eq:3}
    L_2, L_3: y = mx + c \pm{val}
\end{equation}
\begin{equation}\label{eq:4}
    L_4, L_5: y = \frac{-1}{m}x\pm{c}
\end{equation}

In the \textit{first phase}, the rhombus is broken into two subareas (sub rhombuses) to minimize the search space region further. We calculate the number of nodes in the two parts of the bounding area divided by $L_1$, which are $B_{a1}$ surrounded by ($L_1$, $L_2$, $L_4$, and $L_5$) and $B_{a2}$ surrounded by ($L_1$, $L_3$, $L_4$, and $L_5$). We let the new search space $B_{a}$ be the maximum number of nodes between the two subareas, max($num(B_{a1}), num(B_{a2}$)), halving the size of the big rhombus $R_{ba}$. Then, we further halve the size of $B_{a}$ by connecting intersections of ($L_1$, $L_4$), ($L_1$, $L_5$), and the mid point of $L_2$ or $L_3$, creating triangular shaped bounding area, which is 25\% size of $R_{ba}$. This triangular bounding area serves as the initial search space.
    
\begin{figure} [t]
    \centering
    \includegraphics[width=0.6\linewidth]{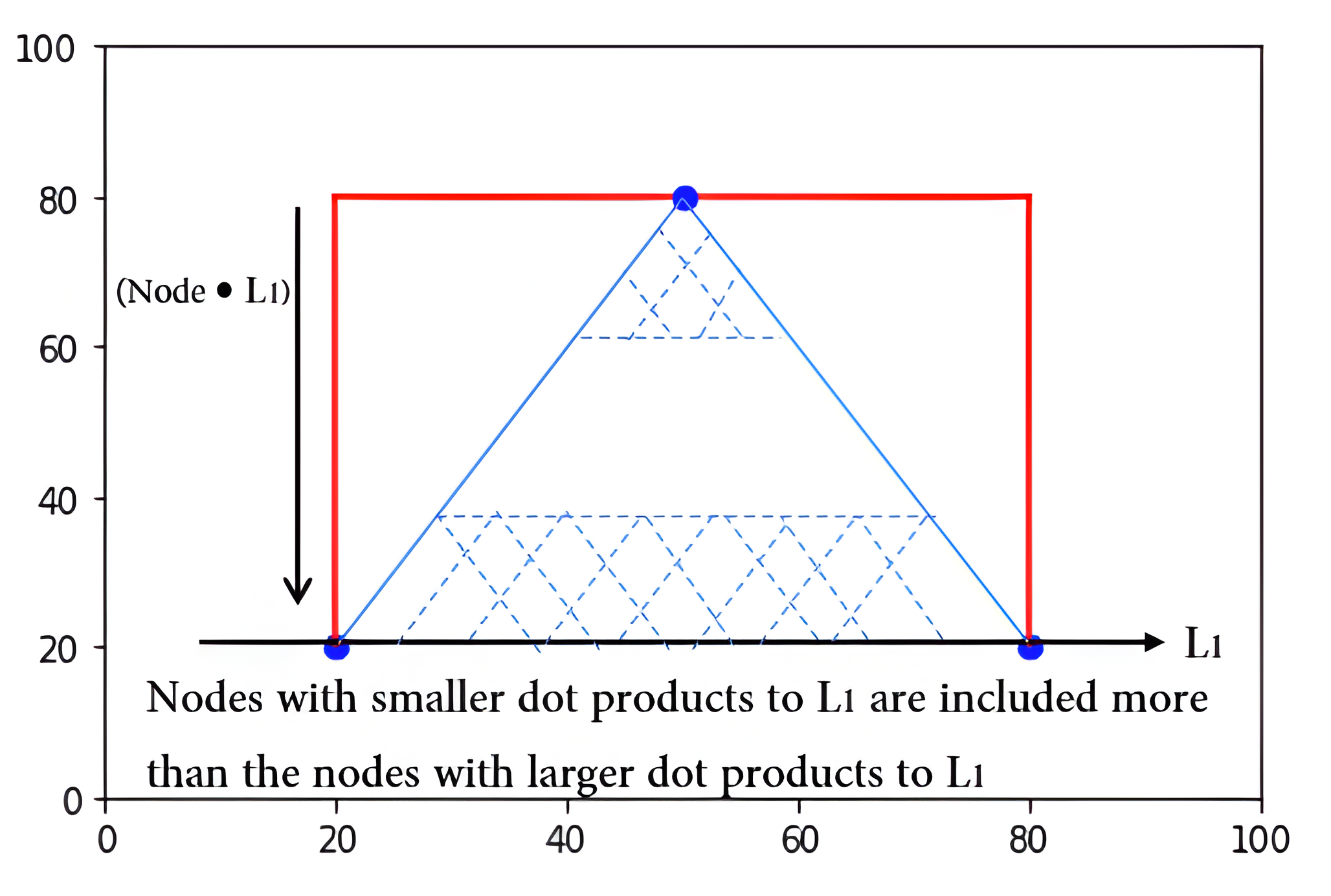}
    \caption{Triangular Bounding Area Illustration}
    \label{Triangle_illustration}
\end{figure}

To describe the intuition behind selecting the triangular subarea to be the initial search space, we use Fig. \ref{Triangle_illustration} to illustrate the concept. Given any node $n$ in the triangle, the distance of the path created connecting A, $n$, and B is shorter than any other node in the sub-rhombus. This is because the dot product of node $n$ and $L_1$ gets smaller inside the triangle. Thus, we give more opportunities to the nodes whose dot products are smaller and located at the bottom of the triangle with a bigger area. As a result, we restrict the opportunities for the nodes located at the top part of the triangle to participate in creating a path. 

The size of the search space $B_{a}$ is incremented in each iteration when a path is not found in four stages: (1) 25\% of $R_{ba}$, (2) 50\% of $R_{ba}$, (3) $R_{ba}$, and (4) Adding closest neighbors of nodes in $R_{ba}$. In the last stage, the number of nodes gets doubled. When given a set of nodes N and a set of edges E, we observe that the computational overhead of consecutive local optimal searches is greater than the computational overhead of incrementing the size of N and E given the search complexity $\mathcal{O}(E + V\log{}V)$. Thus, stage (1) can be \textit{skipped} if the num(nodes) in (1) is less than num(nodes) in $R_{ba} \times {0.25}$. Similarly, stage (2) can be skipped if num(nodes) in (2) is less than num(nodes) in $R_{ba} \times {0.5}$.

The three bounding areas are created in Algorithm \ref{two-phased}. The areas are $R_{ba}$, the rhombus-shaped bounding area, $B_{a3}$, 50\% size of $R_{ba}$, and $B_{a}$, 25\% size of $R_{ba}$ (Lines 2-6). In phase one, a local optimal path search is performed on the current area under the condition that the current bounding area has more nodes than the prescribed number of nodes (Lines 9-16). The algorithm moves into phase two if a path is not found in phase one. In phase two, the algorithm iteratively adds each node's neighbor with the shortest distance in the current bounding area (Lines 18-20). The iteration terminates when a path is found, or a global optimal path search is used.

The time complexities of three algorithms are calculated, where \( V \) represents the number of nodes, \( E \) the number of edges, and \( I \) the number of iterations involved. Let the baseline approach, Dijkstra's algorithm, has a time complexity of \(\mathcal{O}((E+V)\log{V})\).

In the Radius-Based Reactive Service Composition, each iteration adds an \( \mathcal{O}(V) \) complexity when a path is not found within the limited search space, as nodes are added to the next search space. The overall time complexity for this approach is shown in Equation \ref{eq:radius_based} as follows:

\begin{equation}
    \label{eq:radius_based}
    \mathcal{O}(((E+V)\log{V} + {V}) * I)
\end{equation}

Similarly, the Cell Density-Based Service Composition repeats Dijkstra's algorithm in a limited search space until a path is found. This approach requires an additional \( \mathcal{O}(V) \) to compute the density of each cell before running the path-finding algorithm, as shown in Equation \ref{eq:cell_density_based}:

\begin{equation}
    \label{eq:cell_density_based}
    \mathcal{O}({V} + ((E+V)\log{V} + {V}) * I)
\end{equation}

The Two-Phased Reactive Service Composition also uses the Dijkstra algorithm in each iteration within a limited search space when a path is not found. However, after the third iteration, the algorithm only adds the nearest neighbors of existing nodes in the current search space for the next iteration. This reduces the complexity from \( \mathcal{O}(V) \) to \( \mathcal{O}(V_{V\in currentSearchSpace}) \), as indicated in Equation \ref{eq:two_phased}:

\begin{equation} \label{eq:two_phased}
    \mathcal{O}(((E+V)\log{V} + {V}) * I)
\end{equation}
 
These three approaches- radius-based, cell density-based, and the two-phased method- provide distinct strategies for finding the optimal route with minimal search space between disconnected nodes A and B. The radius-based algorithm is advantageous in scenarios requiring a simple search space computation, leveraging the circle radius heuristic. In contrast, the cell density-based algorithm provides a more realistic heuristic by considering the node density around disconnected nodes and adjusting the search space accordingly. The two-phased algorithm prioritizes the region between nodes A and B without considering neighboring nodes extensively. However, it incurs an execution cost, making it less ideal for situations that demand a quick solution.

\section{Performance Evaluation}
We evaluate the performance of our proposed reactive service composition approaches for drone delivery services under the following settings:

\begin{itemize}
    \item \textbf{Performance Metrics:}
    We use (1) \textit{execution time}, (2) \textit{node compression rate}, (3) \textit{line compression rate}, and (4) \textit{distance traveled} as the performance metrics. In this paper, the computational complexity of an algorithm is evaluated using the execution time.
    \item \textbf{Baseline:} We use \textit{Dijkstra}-based service composition approach as a baseline approach to perform local and global drone service compositions.
\end{itemize}

\begin{figure} [t]
    \centering
    \includegraphics[width=0.6\linewidth]{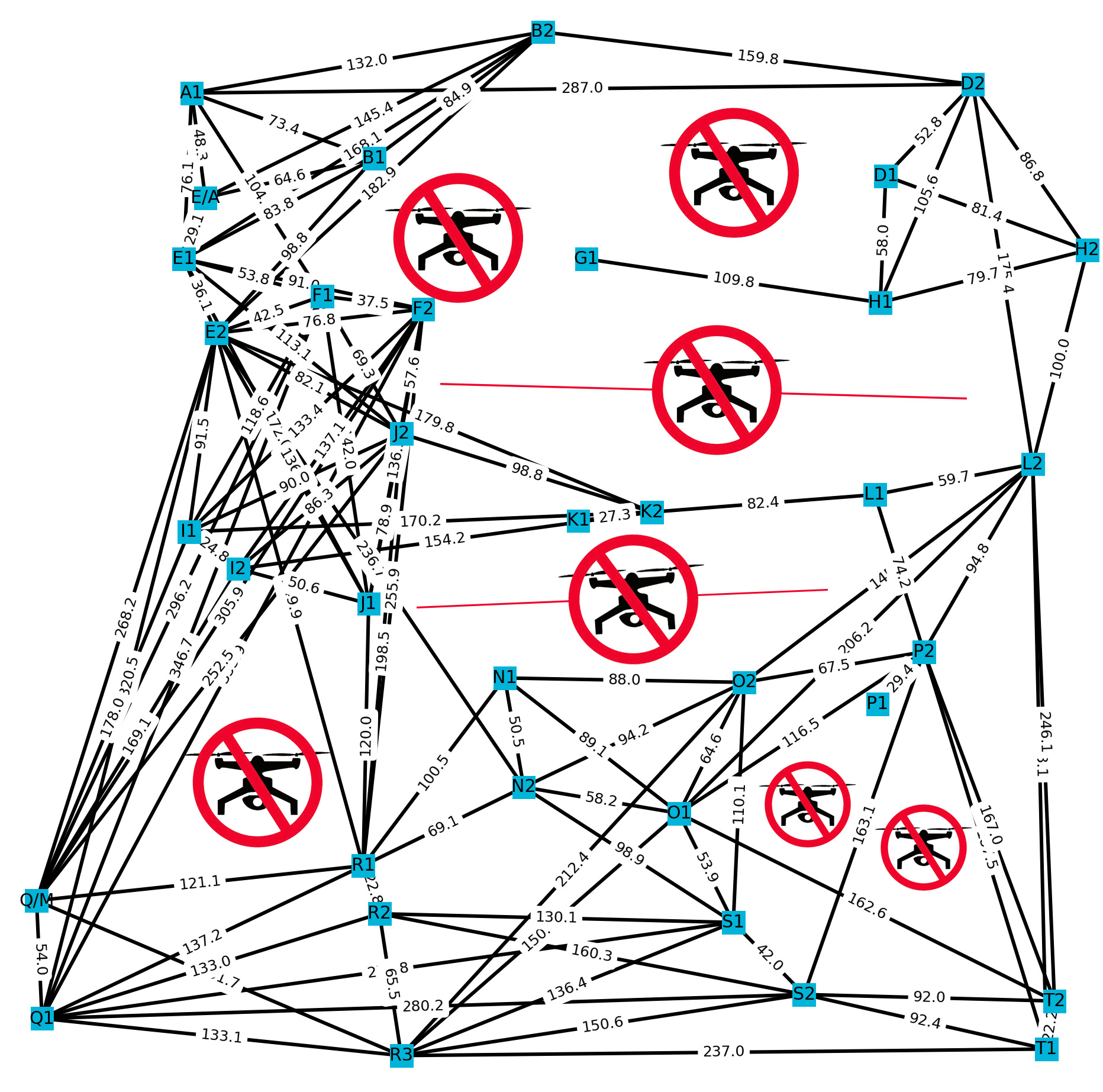}
    \caption{Skyway Network Infrastructure with No-fly Zones}
    \label{skywaynetworkfig}
\end{figure}

\subsection{Experiment Settings with Real Drone Dataset} 

We build a skyway network infrastructure using the NetworkX Python library. Each node in the skyway network represents a delivery target or a recharging station. Each line segment in the skyway network represents a delivery service that is served by a drone. We collect a real drone dataset using an indoor drone testbed for the 3D model of Sydney CBD (Fig. \ref{skywaynetworkfig}). The drone dataset is collected considering the no-fly zones and LOS drone flying regulations. The collected dataset contains data for nodes along with their XY coordinates and the length of each line segment between any two nodes. The real drone dataset consists of a finite number of nodes that are used to construct a small skyway network. We also augment a large skyway network to illustrate the scalability of the proposed approaches. We perform a set of experiments based on the aforementioned performance metrics to evaluate the performance of the proposed approaches. For each experiment, we select a random source and destination node. The experimental variables are described in Table \ref{tab:table1}. The experiments are run on a computer with an Intel i7-3770 processor and 16GB RAM under Windows 10. All the algorithms are written in Python.

\begin{table}
\centering
\caption{Experimental Variables}
\label{tab:table1}
\begin{tabular}{ | m{5cm} | m{7em}| } 
  \hline
  \textbf{Variable} & \textbf{Value} \\ 
  \hline
  Number of nodes & $100 \sim 5000$ \\ 
  \hline
  Max connectivity between nodes & $5 \sim 20$ \\ 
  \hline
  Network size & $1000 \sim 10000$ \\ 
  \hline
  Farthest distance a node can have neighbor at  & 0.05 $\times{map size} \sim $  $0.3 \times{map size}$   \\ 
  \hline
\end{tabular}
\end{table}

\subsection{Results and Discussion}

The proposed approaches perform the reactive service composition of failed drone delivery services for the successful delivery of packages. We first evaluate the radius-based reactive service composition approach compared to the baseline Dijkstra-based service composition approach \cite{behun2022recent}. Then, the cell density-based and two-phased reactive service composition approaches are compared with the baseline approach.

\subsubsection{Radius-Based Reactive Service Composition}

The radius-based reactive service composition approach uses a fixed-sized circular bounding area to reduce the search space. The size of the circle is proportional to the map size. We evaluate this approach based on the average execution time and distance traveled.

\begin{figure}
    \centering
    \begin{minipage}{0.49\textwidth}
        \centering
        \includegraphics[width=\textwidth]{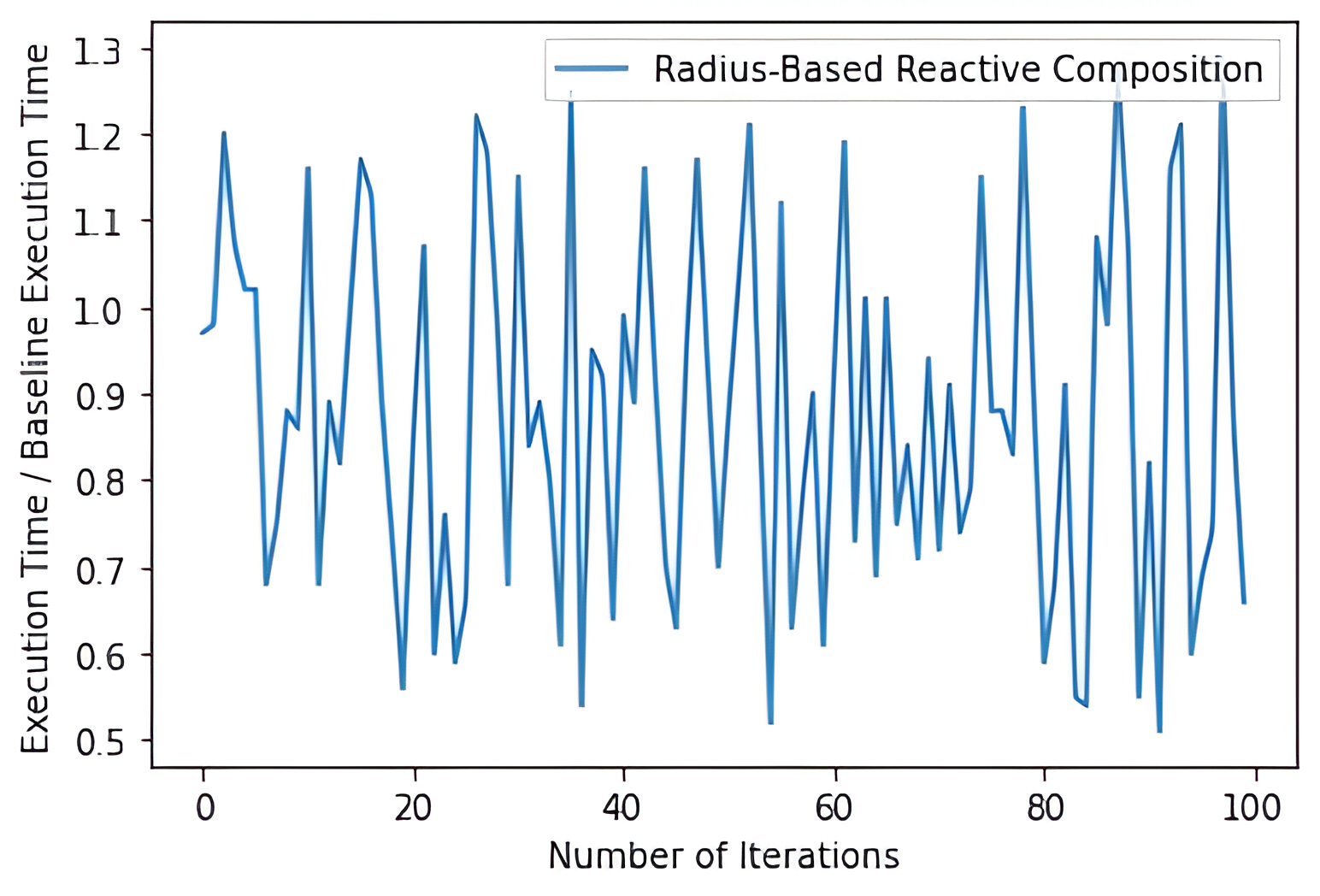} 
        \caption{Average Execution Time}
        \label{first_method}
    \end{minipage}\hfill
    \begin{minipage}{0.49\textwidth}
        \centering
        \includegraphics[width=\textwidth]{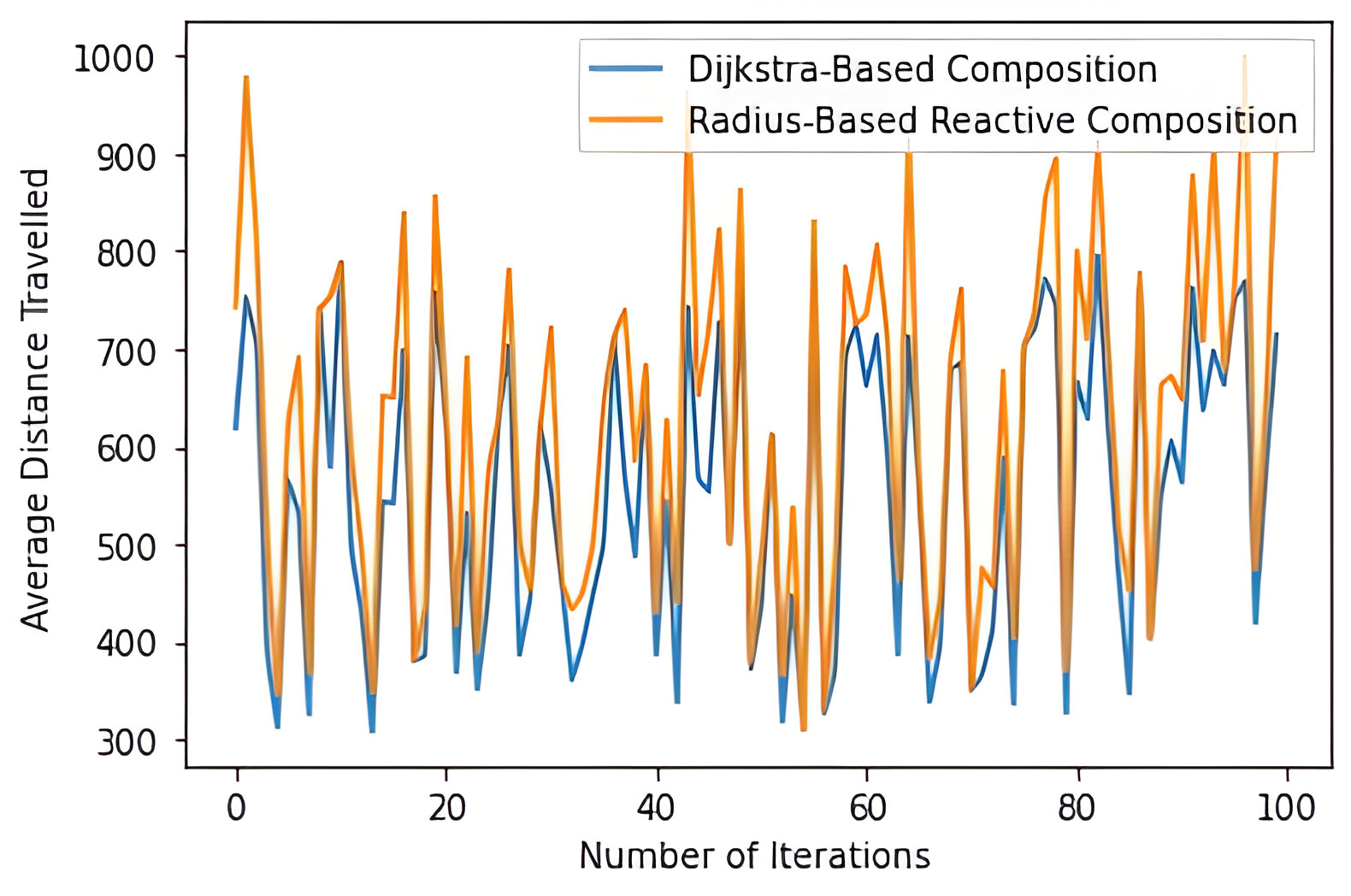} 
        \caption{Average Distance Traveled}
        \label{first_method_dist}
    \end{minipage}
\end{figure}

\textbf{Average Execution Time:} The computational complexity is vital to evaluating the performance of an algorithm. Therefore, we evaluate the computational complexity of the proposed algorithms in terms of execution time. Fig. \ref{first_method} shows the ratio of average execution times of the radius-based approach to the baseline approach for an increasing number of iterations. We observe that the radius-based approach is 9\% faster than the baseline approach. This is because the radius-based reactive service composition approach constrains the search space by limiting the number of candidate drone services. However, suppose we iteratively expand the bounding area of the radius-based approach to include more nodes and line segments. In that case, the overall execution time becomes higher, especially in large skyway networks.

\textbf{Average Distance Traveled:} The cost to deliver a 2 kg package within the 10 km range from the warehouse is estimated to be \$0.1 \cite{5}. We use the distance traveled by a drone as a delivery cost function. Fig. \ref{first_method_dist} presents the average distance traveled by the radius-based and baseline approaches. The results indicate that the radius-based approach travels 14\% more than the baseline approach. This increase in travel distance is due to the restricted search space for finding the optimal services in the radius-based composition approach. In contrast, the baseline approach performs the global composition of services considering all the services operating in the same skyway network. The radius-based reactive service composition approach is not feasible for real-world drone delivery solutions as the distance overhead of this approach is significantly greater than the baseline approach.

\subsubsection{Cell Density-Based and Two-Phased Reactive Service Compositions}

We compare the performance of cell density-based and two-phased composition approaches with the baseline Dijkstra-based composition approach based on the aforementioned performance metrics.

\begin{figure}
    \centering
    \begin{minipage}{0.49\textwidth}
        \centering
        \includegraphics[width=\textwidth]{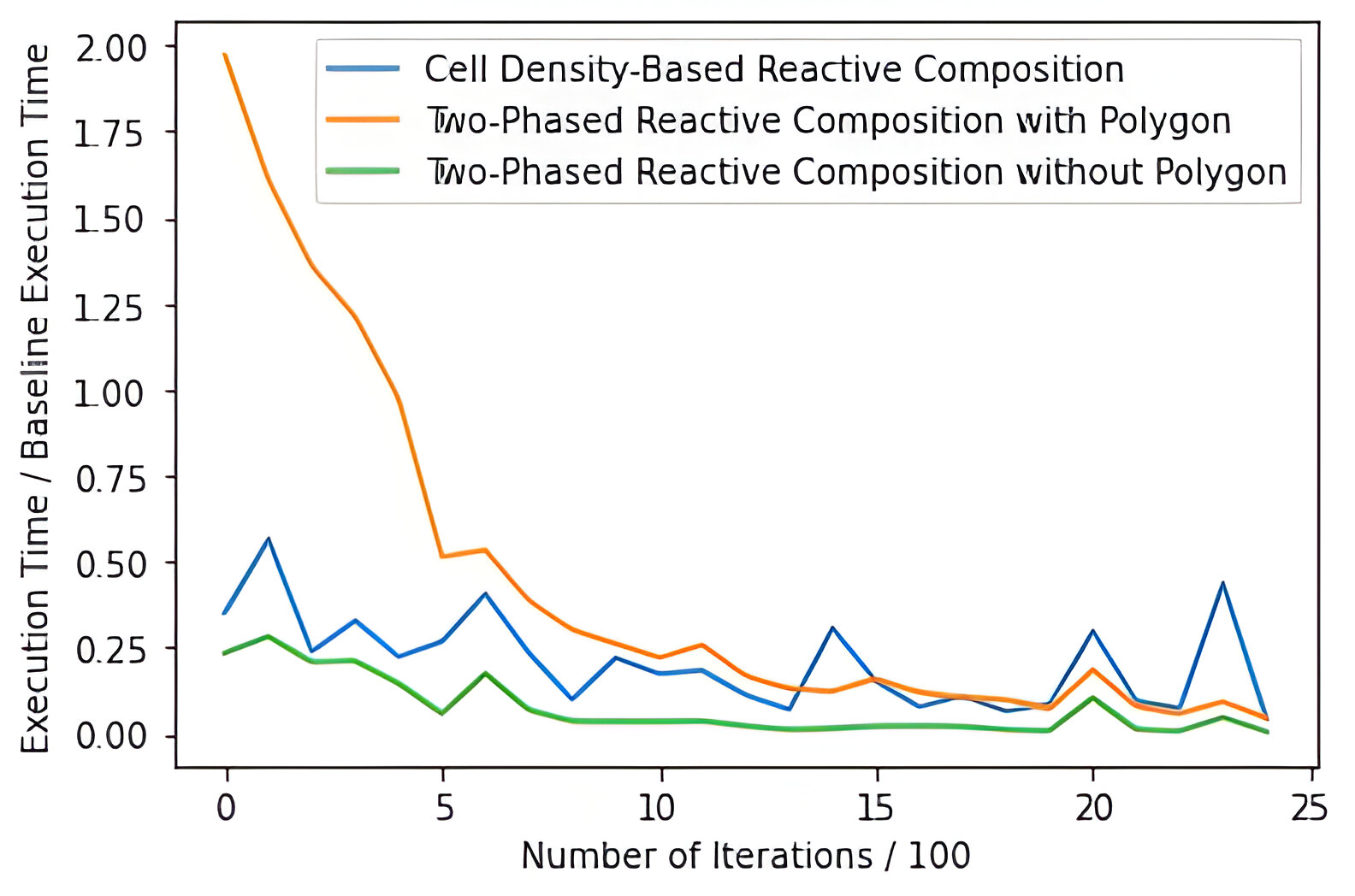} 
        \caption{Average Execution Time}
        \label{time_comp}
    \end{minipage}\hfill
    \begin{minipage}{0.49\textwidth}
        \centering
        \includegraphics[width=\textwidth]{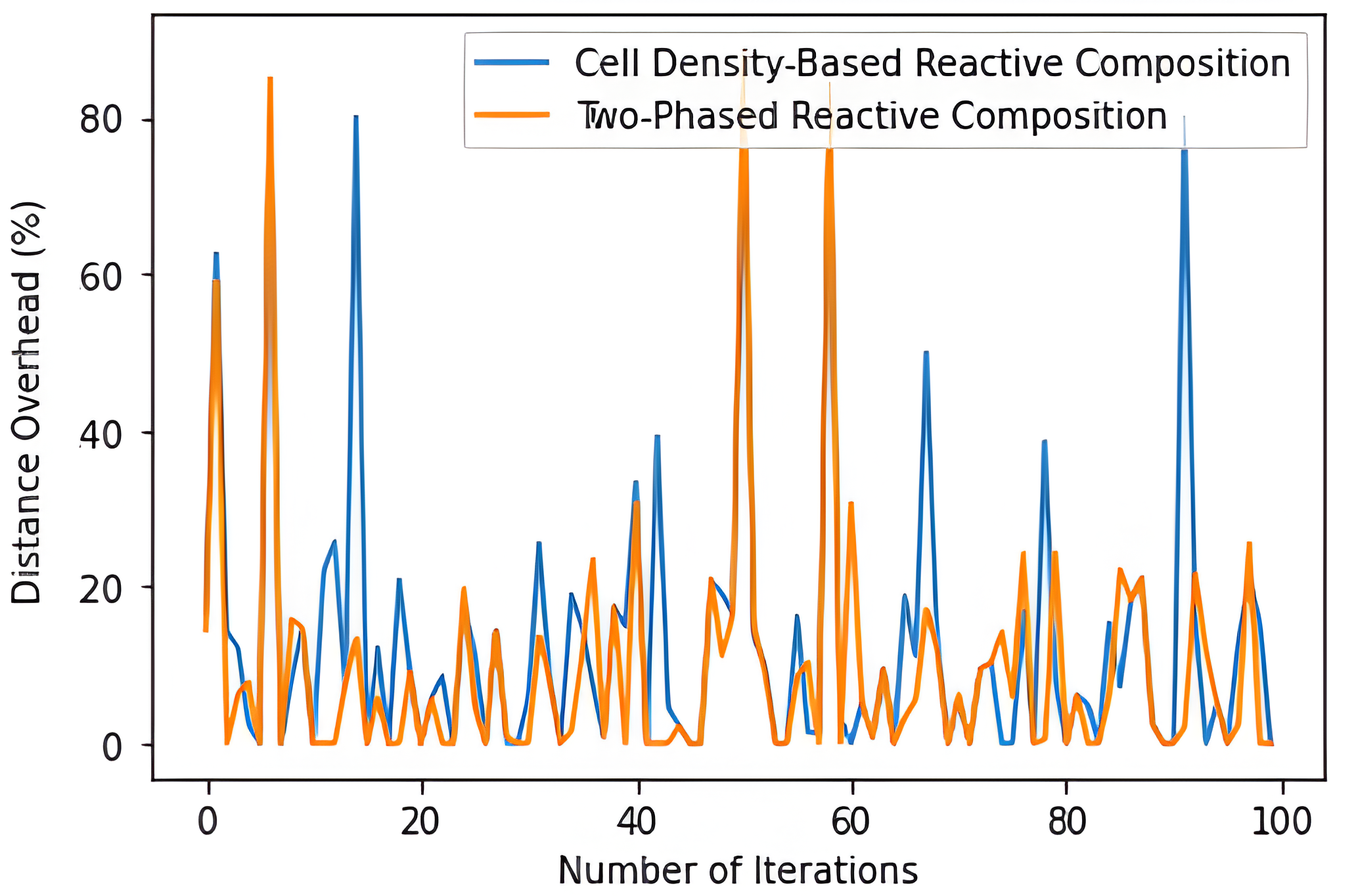} 
        \caption{Distance Overhead}
        \label{dist_comp}
    \end{minipage}
\end{figure}

\textbf{Average Execution Time:} We measure the performance of the two-phased reactive service composition approach in two ways: (1) Execution time with polygons and (2) Execution time without polygons. We gradually increase the skyway network's size and the nodes' densities. Fig. \ref{time_comp} shows the ratio of average execution times of cell density-based and two-phased composition approaches to the baseline approach. The results indicate that the two-phased approach is more computationally efficient than the cell density-based approach. However, the polygon construction is the most time-consuming part of the two-phased reactive service composition approach. The average execution time for the cell density-based approach is 79\% faster, while for the two-phased approach with and without polygon is 60\% and 93\%, respectively.

\textbf{Average Distance Traveled:} Constraining the search space improves the computational efficiency of an algorithm. However, a constrained search space may not provide an optimal delivery solution. Fig. \ref{dist_comp} presents the ratio of travel distance overheads of cell density-based and two-phased reactive service composition approaches to the baseline approach. In this context, travel distance overhead is defined as the total distance traveled by a drone divided by the distance of the baseline approach. The smaller value of travel distance overhead indicates better performance. The results show that the average travel distance overheads using cell density-based and two-phased approaches are 11.21\% and 9.81\%, respectively.

\begin{figure}
    \centering
    \begin{minipage}{0.49\textwidth}
        \centering
        \includegraphics[width=\textwidth]{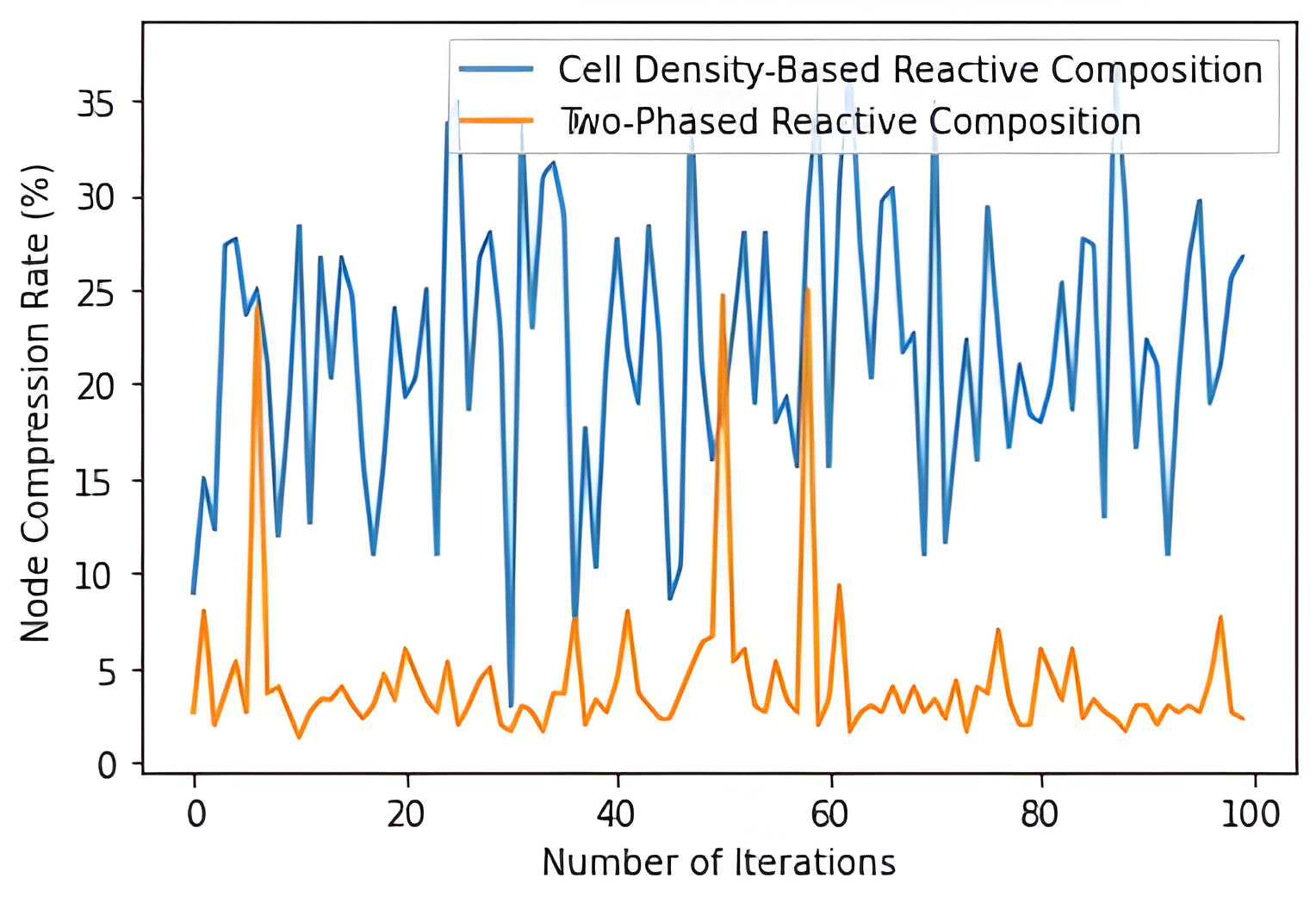} 
        \caption{Node Compression Rate}
        \label{node_comp}
    \end{minipage}\hfill
    \begin{minipage}{0.49\textwidth}
        \centering
        \includegraphics[width=\textwidth]{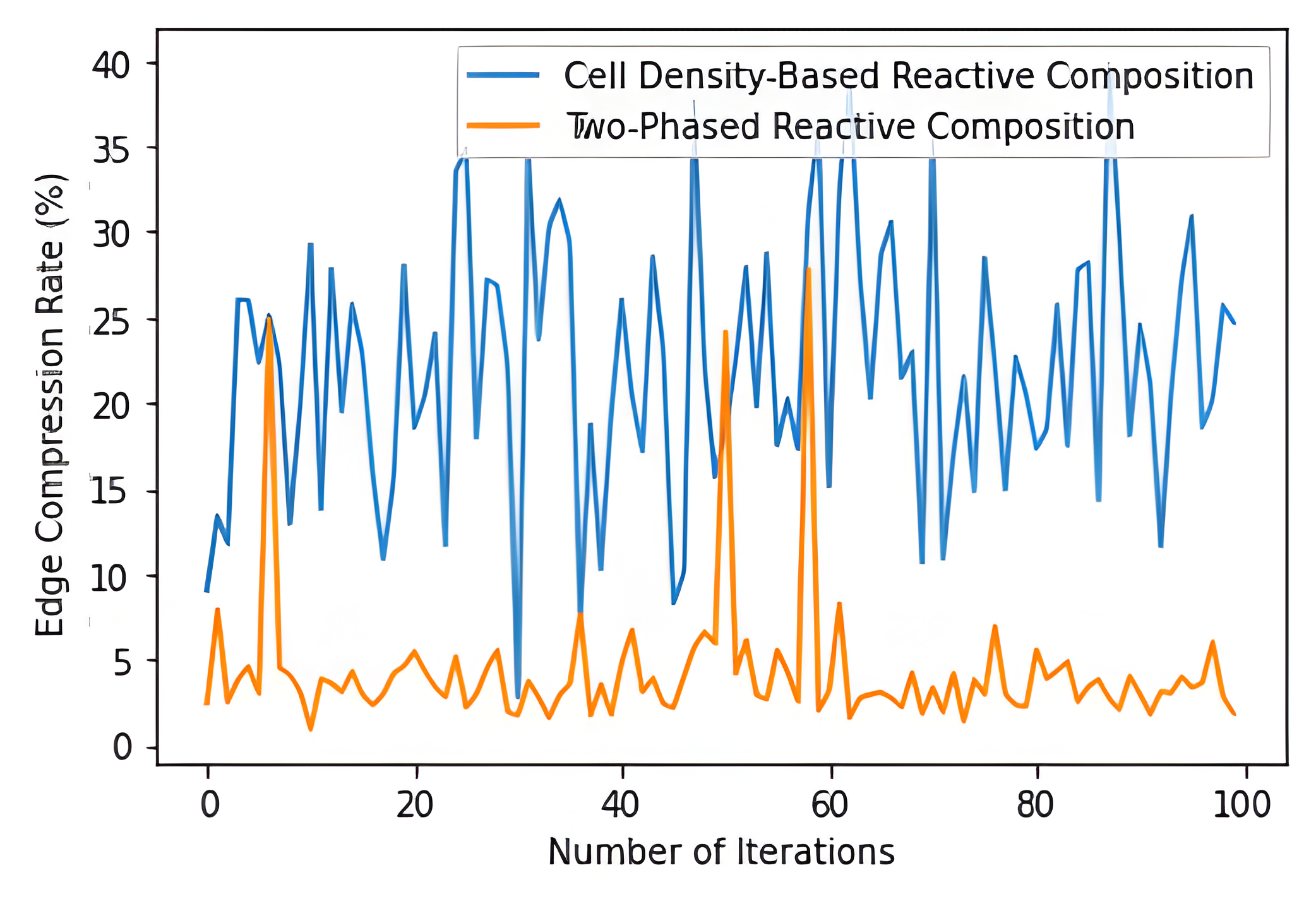} 
        \caption{Edge Compression Rate}
        \label{line_comp}
    \end{minipage}
\end{figure}

\textbf{Node/Edge Compression Rate:} We investigate a skyway network's node/edge compression rate that an algorithm requires to enhance the local optimal search. The compression rate quantifies the reduction in the number of nodes within the search space across various algorithms. The node/edge compression rates of cell density-based and two-phased reactive service composition approaches are shown in Fig. \ref{node_comp} and \ref{line_comp}. We observe that the average number of nodes/edges for cell density-based and two-phased approaches are 20\% and 4\% of the original skyway network, respectively. Furthermore, the two-phased approach outperforms the cell density-based approach as it computes an optimal composition plan using a significantly smaller number of nodes/edges.

\begin{figure}
    \centering
    \begin{minipage}{0.49\textwidth}
        \centering
        \includegraphics[width=\textwidth]{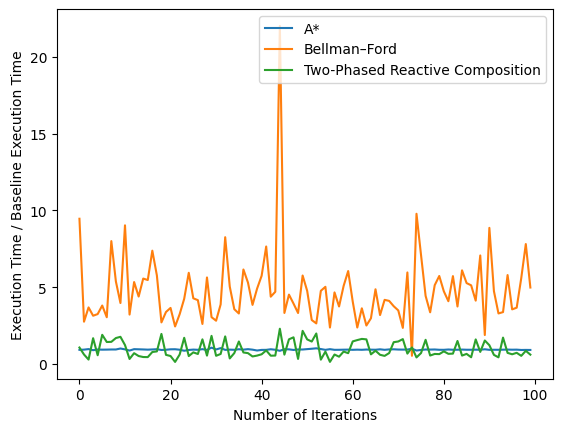} 
        \caption{Average Execution Time}
        \label{Time_Comparison_3_algorithms}
    \end{minipage}\hfill
    \begin{minipage}{0.49\textwidth}
        \centering
        \includegraphics[width=\textwidth]{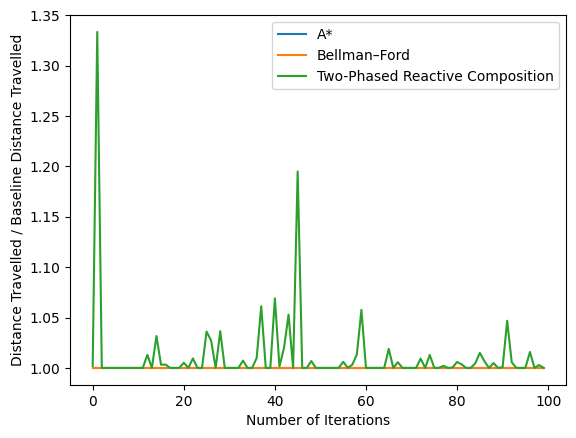} 
        \caption{Average Distance Traveled}
        \label{Dist_Comparison_3_Algorithms}
    \end{minipage}
\end{figure}

\subsubsection{Two-Phased Reactive Service Composition} 

We compare the best-performing algorithm among the three proposed algorithms, Two-Phased Reactive Service Compositions, with the baseline approach (Dijkstra), the A* algorithm \cite{A*Pathfinding} with Euclidean distance between the current node and the destination as the heuristic, and the Bellman-Ford algorithm \cite{7287776}.

\textbf{Average Execution Time:} Fig. \ref{Time_Comparison_3_algorithms} presents the average execution times for all three algorithms normalized against the baseline approach. We observe that the A* algorithm consistently performs close to the baseline, taking only 6.5\% less time on average. The Two-Phased Reactive Composition demonstrates significant efficiency, outperforming the baseline by 16\% on average, with its execution time well below the baseline throughout the iterations. However, the Bellman-Ford algorithm takes a substantially longer execution time, with an average of nearly five times the baseline. This high variability and longer execution time show that Bellman-Ford is less efficient than the other algorithms.

\textbf{Average Distance Traveled:} Fig. \ref{Dist_Comparison_3_Algorithms} depicts the average distance traveled for three algorithms normalized against the baseline approach. We observe that the A* and Bellman-Ford algorithms maintain the same performance as the baseline approach. However, the Two-Phased Reactive Composition algorithm incurs a slight overhead on average. The distance traveled by the Two-Phased Reactive Composition algorithm is 1.2\% longer than the other algorithms.

Overall, the Two-Phased Reactive Service Composition outperforms the baseline and Bellman-Ford approaches in execution time. There is a slight overhead in the distance traveled, but it is nearly the same as the A* algorithm.

\subsubsection{Effectiveness of Stage Skipping in Two-Phased Reactive Service Composition}

We use the stage skipping strategy for 25\% and 50\% size bounding areas of $R_{ba}$. The stage skipping is performed when the number of nodes in $R_{ba}$ is less than the appropriate percentage size of $R_{ba}$. For example, we perform stage skipping for 25\% of $R_{ba}$ if there does not exist at least 25\% nodes in $R_{ba}$.

\begin{figure}
    \centering
    \begin{minipage}{0.49\textwidth}
        \centering
        \includegraphics[width=\textwidth]{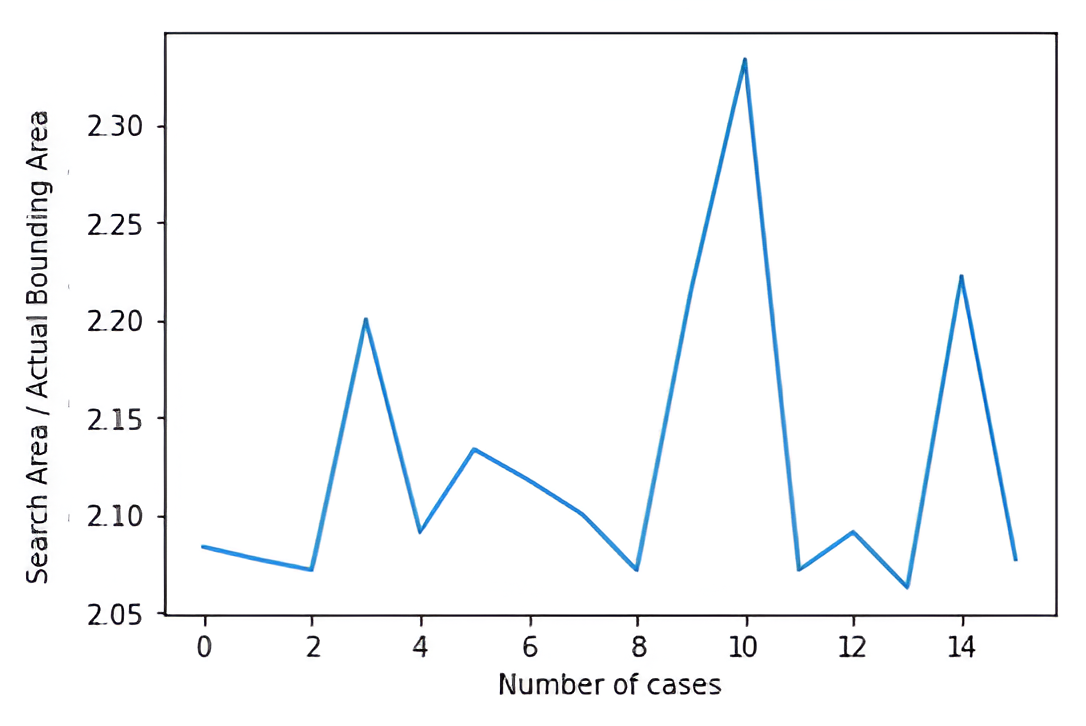} 
        \caption{Ineffective Skipping}
        \label{skipped_1}
    \end{minipage}\hfill
    \begin{minipage}{0.49\textwidth}
        \centering
        \includegraphics[width=\textwidth]{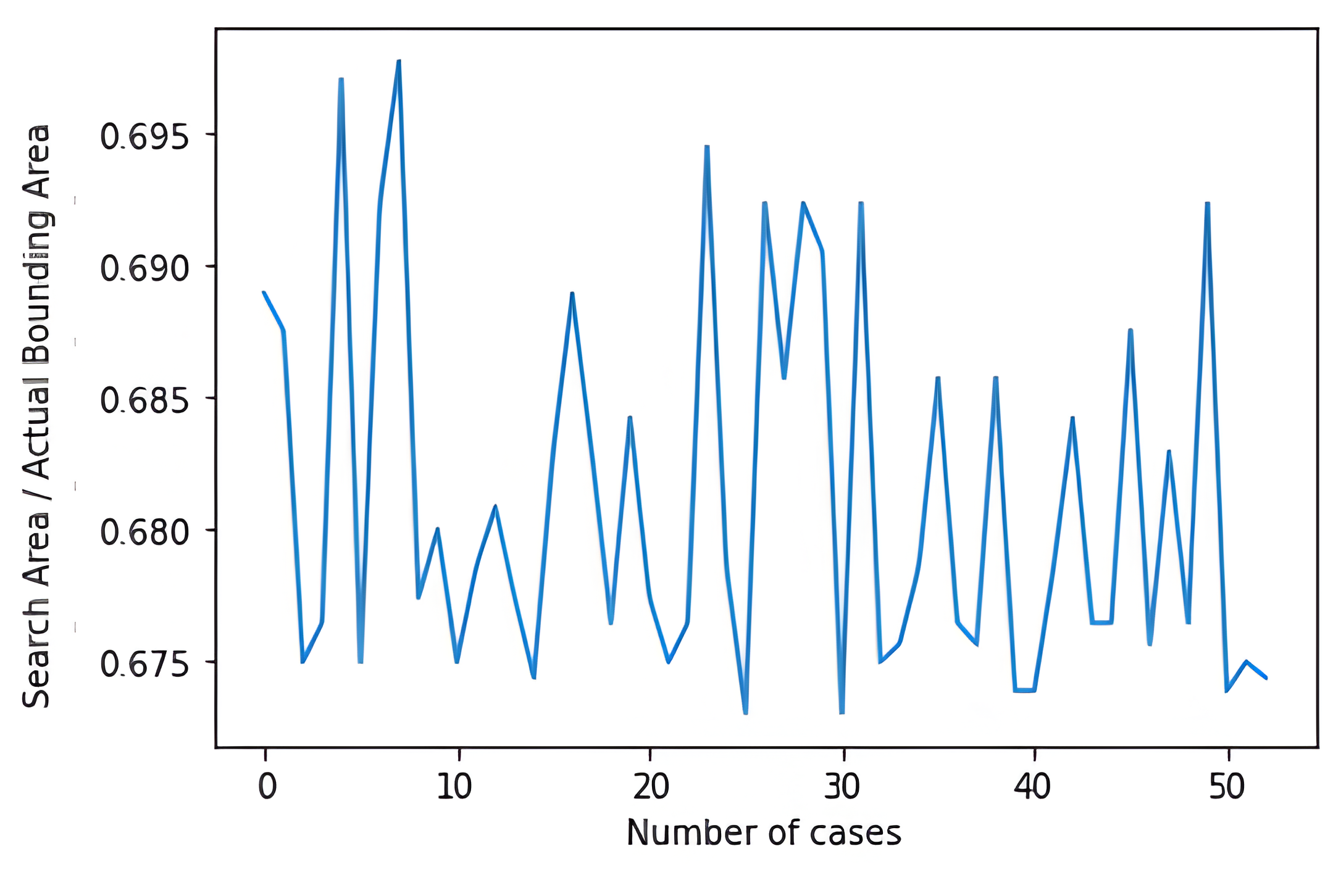} 
        \caption{Effective Skipping}
        \label{skipped_2}
    \end{minipage}
\end{figure}

\begin{figure}
    \centering
    \begin{minipage}{0.49\textwidth}
        \centering
        \includegraphics[width=\textwidth]{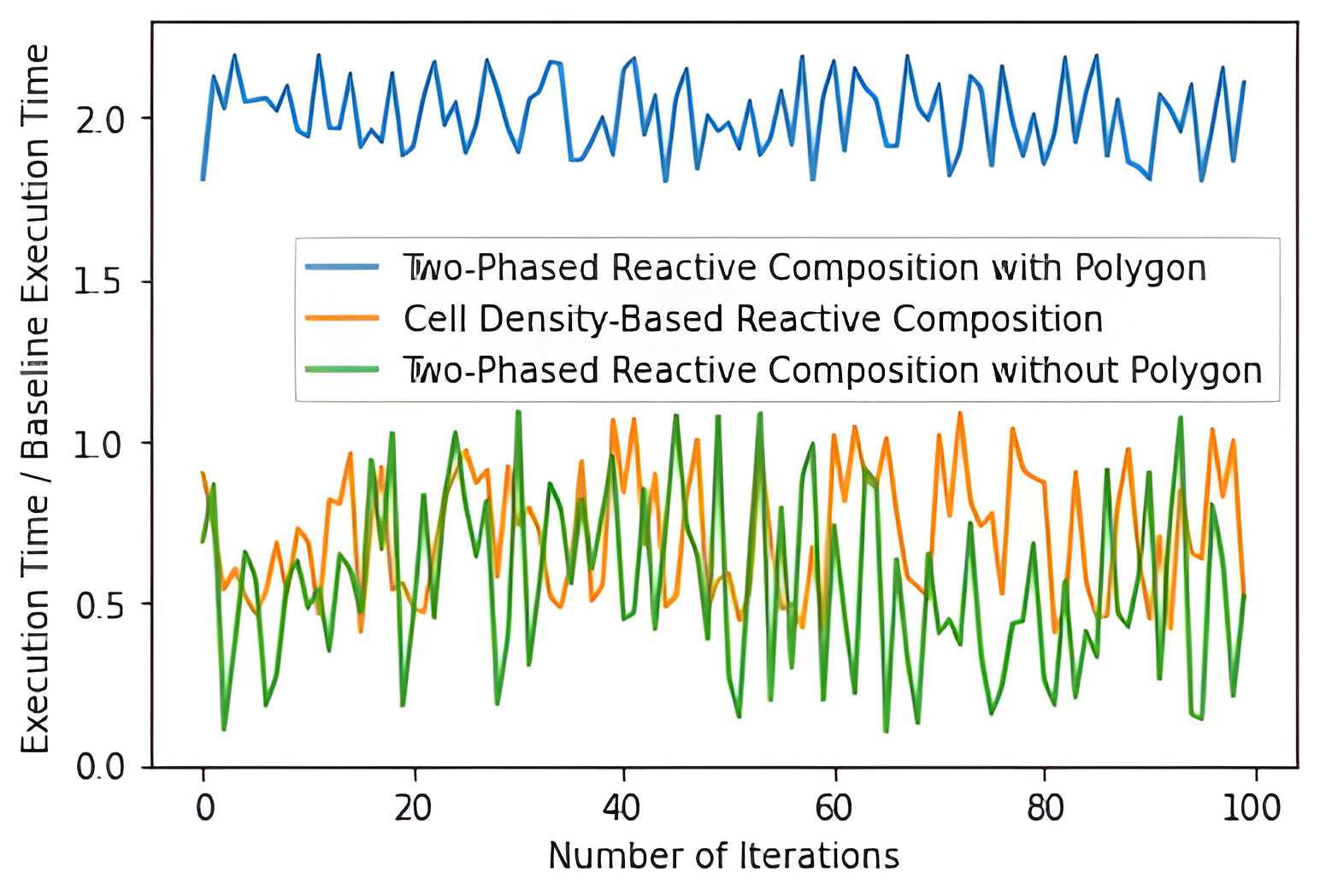} 
        \caption{Average Execution Time}
        \label{real_result}
    \end{minipage}\hfill
    \begin{minipage}{0.49\textwidth}
        \centering
        \includegraphics[width=\textwidth]{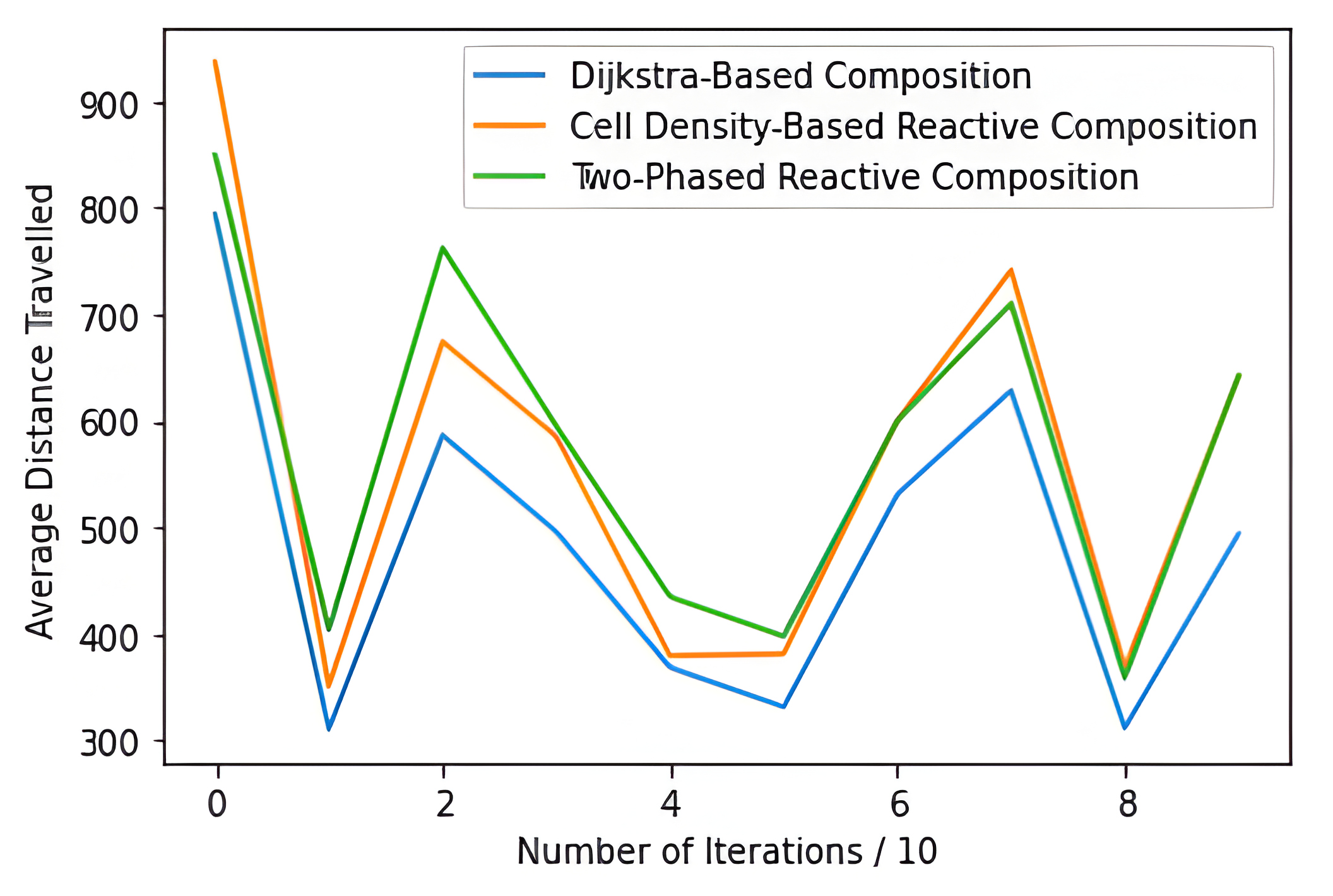} 
        \caption{Average Distance Traveled}
        \label{real_result_dist}
    \end{minipage}
\end{figure}

A local optimal service composition may exist without following the stage skipping strategy. Fig. \ref{skipped_1} shows the ineffective skipping where the local optimal compositions exist in the skipped stages. The number of iterations to compute the compositions remains the same. However, the number of nodes/edges increases when the bounding area expands by skipping the previous stage. The results show that the average search space required to compute a local optimal composition in case of ineffective skipping is 2.13 times larger than without stage skipping. Fig. \ref{skipped_2} shows the effective skipping where the skipped stage does not contain a local optimal composition plan. The benefits of effective skipping are two-fold: (1) search space is 32\% smaller than the space required by ineffective skipping, and (2) the number of iterations required to compute a local optimal composition becomes significantly less by skipping the stages. The results indicate that the stage skipping strategy is effective in 53 cases while ineffective in 16 cases out of 69 cases.

\subsubsection{Evaluation in Sydney CBD Skyway Network} We evaluate the effectiveness of cell density-based and two-phased approaches in the Sydney CBD skyway network (Fig. \ref{skywaynetworkfig}). This evaluation demonstrates the usefulness of our proposed approaches in real-world scenarios for the successful provisioning of drone delivery services. Fig. \ref{real_result} presents the ratio of average execution times of cell density-based and two-phased composition with and without polygon approaches to the baseline approach. The results indicate that the two-phased composition approach with a polygon consumes twice as much time as the baseline approach. On the other hand, cell density-based and two-phased without polygon approaches require 56\% and 74\% less execution time than the baseline approach. We observe a clear trade-off between the average execution time and the average distance traveled by a drone for these approaches. This observation is based on results in Fig. \ref{real_result_dist}, where the drone travels 14\% and 11\% more for cell density-based and two-phased composition approaches compared to the baseline approach, respectively.

\section{Conclusion}

We proposed a novel failure-aware reactive UAV delivery service composition framework. The proposed framework included a skyway network infrastructure, a formal drone delivery service model, and a system architecture for reactive drone delivery services. The framework also comprised reactive service composition algorithms for drone delivery services using radius-based, cell density-based, and two-phased approaches. We used these approaches to constrain the search space and performed reactive service compositions when a service failure occurred. We conducted a set of experiments with a real drone dataset to illustrate the performance of our proposed approaches. We observed that radius-based reactive service composition was not a practical solution for real-world scenarios. Compared to the baseline approach for the increasing skyway network size, its performance gradually decreased. We also observed that our proposed approaches are computationally efficient compared to the baseline approach. The two-phased approach provided overall better solutions than the cell density-based and baseline approaches. In the future, we plan to investigate the weather patterns, such as wind and temperature, that may cause QoS fluctuations and failures in drone delivery services.

\section*{Acknowledgment}

This research was partly made possible by LE220100078 and DP220101823 grants from the Australian Research Council. The statements made herein are solely the responsibility of the authors.

\bibliographystyle{IEEEtran}
\bibliography{references}

\begin{IEEEbiography}[{\includegraphics[width=1in,height=1.25in,clip,keepaspectratio]{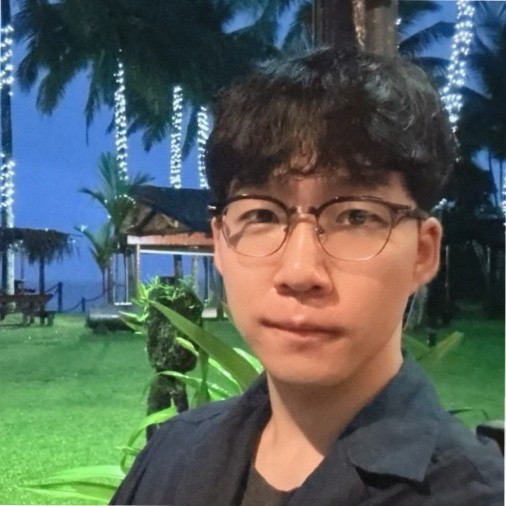}}]{Woojin Lee}{\,}is a Software Engineer at WiseTech Global. He completed his Honours degree under the supervision of Prof. Athman Bouguettaya at the University of Sydney. His interests focus on Drone-based Services in smart cities.
\end{IEEEbiography}

\begin{IEEEbiography}[{\includegraphics[width=1in,height=1.25in,clip,keepaspectratio]{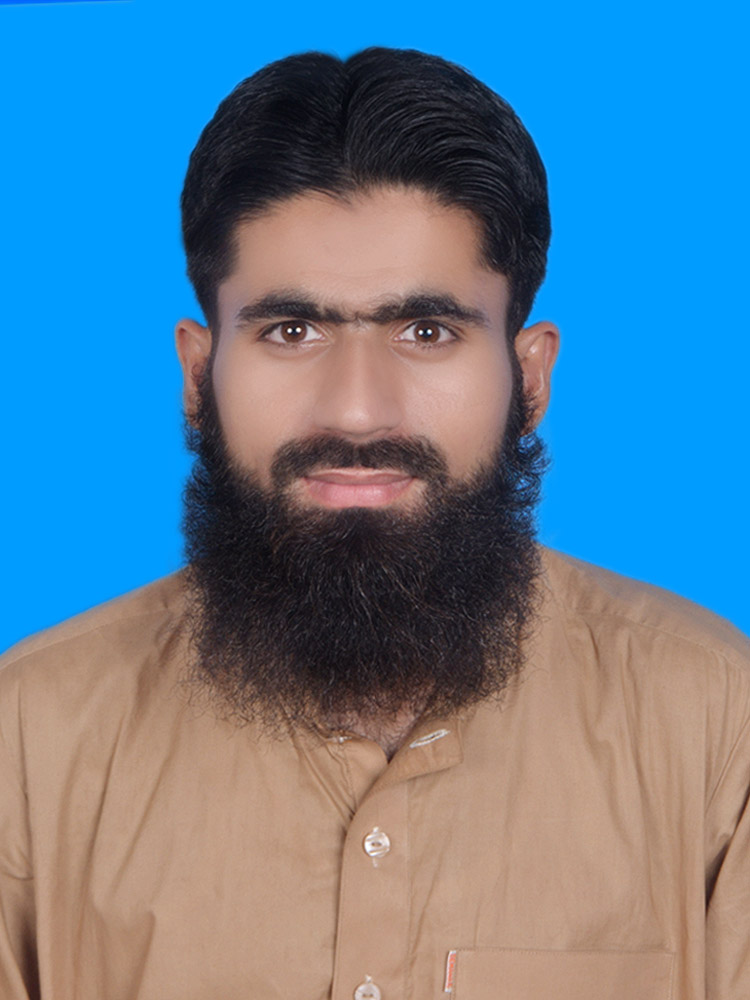}}]{Babar Shahzaad}{\,}is a Research Fellow in the School of Information Systems at Queensland University of Technology (QUT). He received his Ph.D. in Computer Science from the University of Sydney. He has published in top-ranked conferences and journals, including IEEE ICWS, ICSOC, IEEE IoT, and FGCS. His research interests include the Industrial Internet of Things (IIoT), ICN/NDN applications for IoT, Service Computing, and Drone-based Delivery Services in Smart Cities.
\end{IEEEbiography}

\begin{IEEEbiography}[{\includegraphics[width=1in,height=1.25in,clip,keepaspectratio]{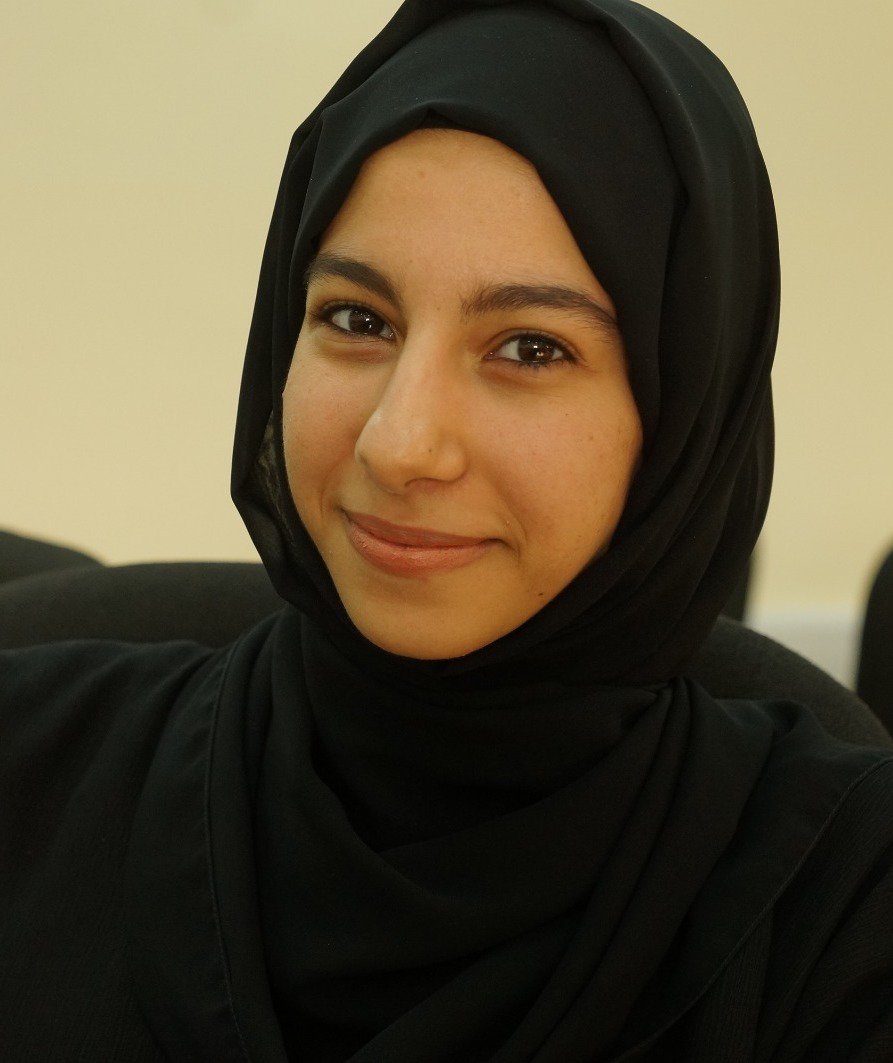}}]{Balsam Alkouz}{\,}is a Postdoctoral Fellow at the School of Computer Science at the University of Sydney. She received her Ph.D. degree in computer science from the University of Sydney. Australia, in 2023. Balsam completed her bachelor's degree in IT Multimedia and her master’s degree in Computer Science from the University of Sharjah, United Arab Emirates, in 2016 and 2018 respectively. She worked as a Research Assistant in the Data Mining and Multimedia Research Group at the University of Sharjah. Her research focuses on IoT, Service Computing, and Data Mining.
\end{IEEEbiography}

\begin{IEEEbiography}[{\includegraphics[width=1in,height=1.25in,clip,keepaspectratio]{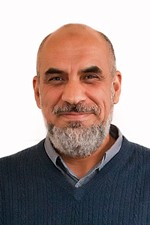}}]{Athman Bouguettaya}{\,}is a Professor in the School of Computer Science at the University of Sydney. He received his Ph.D. in Computer Science from the University of Colorado at Boulder (USA) in 1992. He is or has been on the editorial boards of several journals, including the IEEE Transactions on Services Computing, ACM Transactions on Internet Technology, the International Journal on Next Generation Computing, and VLDB Journal. He is a Fellow of the IEEE and a Distinguished Scientist of the ACM. He is a member of the Academia Europaea (MAE).
\end{IEEEbiography}

\vfill

\end{document}